\documentclass[10pt,twocolumn,letterpaper]{article}
 
\usepackage{iccv}
\iccvfinalcopy
\usepackage{times}

\usepackage{epsfig}
\usepackage{graphicx}
\usepackage{amsmath}
\usepackage{amssymb}
\usepackage{mathtools}
\usepackage{array}
\usepackage{soul}
\newcolumntype{P}[1]{>{\centering\arraybackslash}p{#1}}
\usepackage{float}
\usepackage{pifont}%
\usepackage{multirow}
\usepackage{nicefrac}       %
\usepackage{microtype}  
\usepackage{tabularx}
\usepackage{array}
\usepackage{enumitem}
\usepackage{wrapfig,booktabs}

\usepackage{afterpage}
\usepackage{scalerel,xparse}
\usepackage{capt-of,etoolbox}
\usepackage{amsmath}
\usepackage{amsfonts}
\usepackage{url}            %
\usepackage{booktabs}       %
\usepackage{amsfonts}       %
\usepackage{nicefrac}       %
\usepackage{microtype}      %
\usepackage{xcolor}         %
\usepackage{graphics}
\usepackage{xfrac}
\usepackage{colortbl}\newcommand{\vect}[1]{{{\bf #1}}}
\usepackage[pagebackref,breaklinks,colorlinks]{hyperref}
\usepackage[capitalize]{cleveref}
\crefname{section}{Sec.}{Secs.}
\Crefname{section}{Section}{Sections}
\Crefname{table}{Table}{Tables}
\crefname{table}{Tab.}{Tabs.}

\newcommand{\dafmax}[1]{{\begin{array}{c}\mbox{max}\\{#1}\end{array}}}

\newcolumntype{N}{>{\centering\arraybackslash\footnotesize}m{.5in}}
\newcolumntype{G}{>{\centering\arraybackslash}m{34pt}}
\usepackage[pagebackref,breaklinks,colorlinks]{hyperref}
\usepackage[capitalize]{cleveref}
\crefname{section}{Sec.}{Secs.}
\Crefname{section}{Section}{Sections}
\Crefname{table}{Table}{Tables}
\crefname{table}{Tab.}{Tabs.}

\def\iccvPaperID{8591} %

\ificcvfinal\fi
\begin{document}
\title{StyLitGAN: Prompting StyleGAN to Produce New Illumination Conditions
}

\author{%
{Anand Bhattad\hspace{1cm} D.A. Forsyth} \\
 University of Illinois Urbana-Champaign\\
  \texttt{\href{https://anandbhattad.github.io/stylitgan/}{https://anandbhattad.github.io/stylitgan/}} \\
}

\makeatletter
\g@addto@macro\@maketitle{
	\begin{figure}[H]
	\scriptsize
		\setlength{\linewidth}{\textwidth}
		\setlength{\hsize}{\textwidth}
		\vspace{-7.5mm}
  \centering
  \footnotesize
  \setlength\tabcolsep{0.2pt}
  \renewcommand{\arraystretch}{0.1}

  \begin{tabular}{cccccc}
        Generated Image & Relit - 1 ($\vect{w}^+ + d_1$) &  Relit - 2  ($\vect{w}^+ + d_2$) & Relit - 3  ($\vect{w}^+ + d_3$) & Relit - 4   ($\vect{w}^+ + d_4$)& Relit - 5  ($\vect{w}^+ + d_5$) \\ 
    \includegraphics[width=0.16\linewidth]{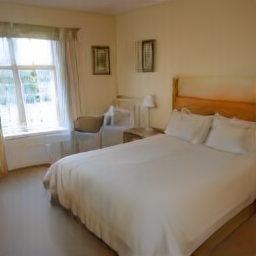} &
    \includegraphics[width=0.16\linewidth]{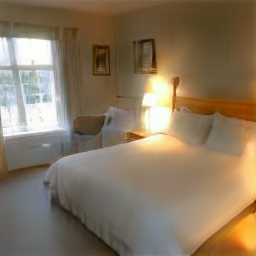} &
    \includegraphics[width=0.16\linewidth]{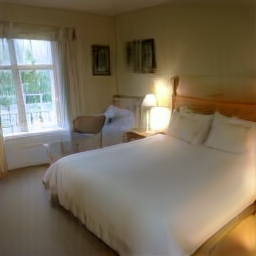} &
    \includegraphics[width=0.16\linewidth]{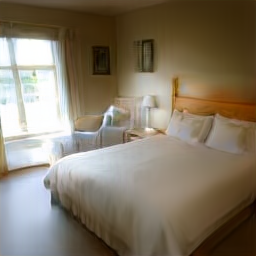} &
    \includegraphics[width=0.16\linewidth]{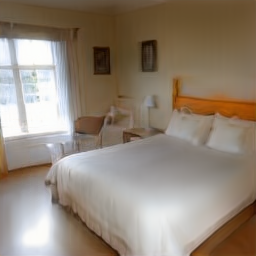} &
    \includegraphics[width=0.16\linewidth]{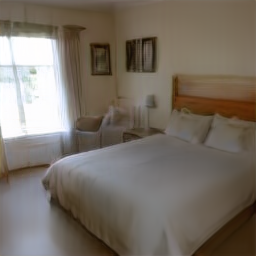} \\
    \includegraphics[width=0.16\linewidth]{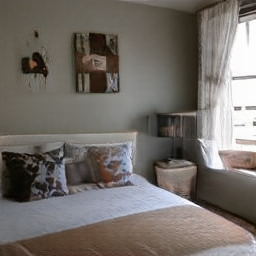} &
    \includegraphics[width=0.16\linewidth]{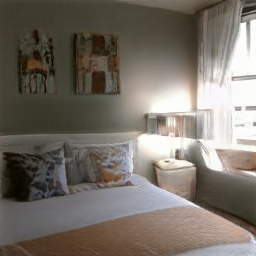} &
    \includegraphics[width=0.16\linewidth]{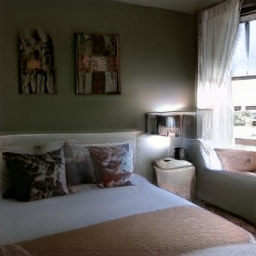} &
    \includegraphics[width=0.16\linewidth]{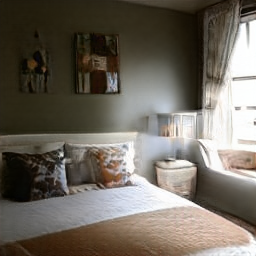} &
    \includegraphics[width=0.16\linewidth]{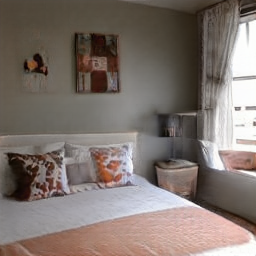} &
    \includegraphics[width=0.16\linewidth]{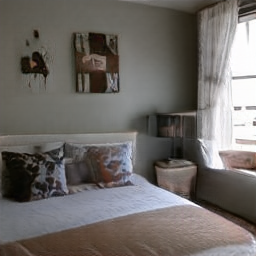} 
            \\
    \includegraphics[width=0.16\linewidth]{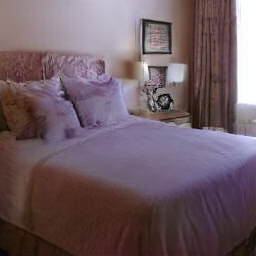} &
    \includegraphics[width=0.16\linewidth]{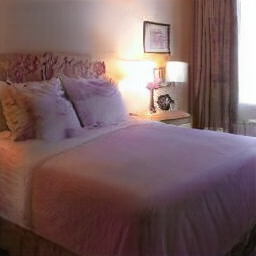} &
    \includegraphics[width=0.16\linewidth]{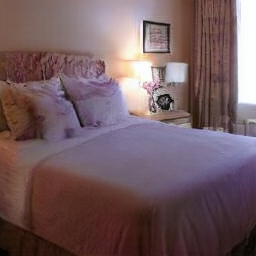} &
    \includegraphics[width=0.16\linewidth]{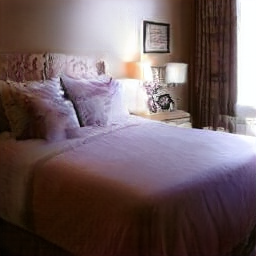} &
    \includegraphics[width=0.16\linewidth]{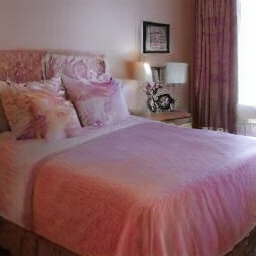} &
    \includegraphics[width=0.16\linewidth]{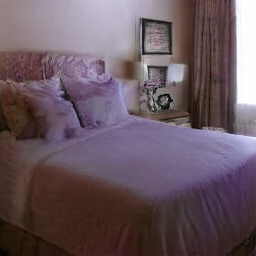} 
    \\
            \includegraphics[width=0.16\linewidth]{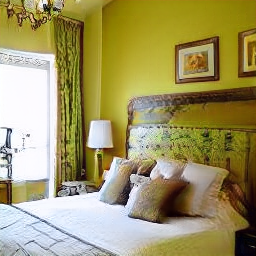} &
    \includegraphics[width=0.16\linewidth]{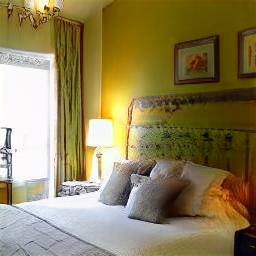} &
    \includegraphics[width=0.16\linewidth]{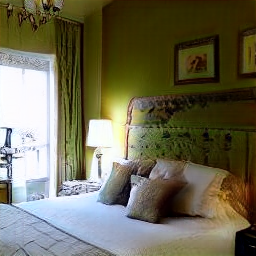} &
    \includegraphics[width=0.16\linewidth]{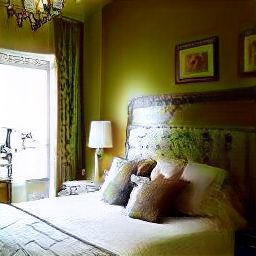} &
    \includegraphics[width=0.16\linewidth]{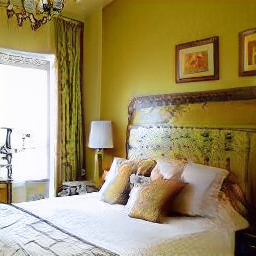} &
    \includegraphics[width=0.16\linewidth]{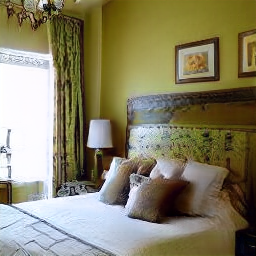} 
        \end{tabular}
      \caption{Our approach, StyLitGAN, searches for a set of directions ($d_i$) in StyleGAN's $\mathcal{W}$ style space that when added to $\vect{w}^+$ style code can change a generated image's lighting while maintaining its albedo. Our method does not require a per-image search or fine-tuning. The first column shows images generated by a vanilla StyleGAN2, while the other columns show the same scenes relit with the same applied direction. The relighting directions ($d_i$) are obtained from a forward selection process to promote diversity in relighting and are not cherry-picked. The illumination of the scene changes significantly, while the geometry and albedo remain unchanged, resulting in realistic images. It is also worth noting that the effects of the directions are consistent across scenes.}
        \label{fig:teaser}
        \vspace{-5pt}
	\end{figure}
}
\makeatother    
\maketitle
\begin{abstract}
We propose a novel method, StyLitGAN, for relighting and resurfacing generated images in the absence of labeled data. Our approach generates images with realistic lighting effects, including cast shadows, soft shadows, inter-reflections, and glossy effects, without the need for paired or CGI data. 

StyLitGAN uses an intrinsic image method to decompose an image, followed by a search of the latent space of a pre-trained StyleGAN to identify a set of directions. By prompting the model to fix one component (e.g., albedo) and vary another (e.g., shading), we generate relighted images by adding the identified directions to the latent style codes. Quantitative metrics of change in albedo and lighting diversity allow us to choose effective directions using a forward selection process. Qualitative evaluation confirms the effectiveness of our method.
\vspace{-5pt}
\end{abstract}
\section{Introduction}
The same scene viewed by the same camera will produce different images as the sun moves in the sky or people turn on or off lights. Similarly, changing the paint color on a wall
will change the pixel values corresponding to the wall and also affect the rest of the image due to changes in the reflection of light from that wall.
Current generative models can produce impressively realistic images from random vectors~\cite{goodfellow2014generative, karras2019style, 
  Karras2019stylegan2, karras2021alias}, but cannot be effectively prompted to alter intrinsic properties such as the lighting in a scene.   In this work, we extend the capabilities of current editing
methods~\cite{voynov2020unsupervised, wu2021stylespace,   shen2020interfacegan, zhu2020indomain} by demonstrating how to fix an image component while changing another.
This allows us to keep the scene geometry fixed and alter the lighting or keep the lighting fixed and alter the color of the surfaces.

Our method, StyLitGAN, is built on StyleGAN ~\cite{
  Karras2019stylegan2} and leverages established procedures to manipulate style codes effectively to edit images. We use StyleGAN to produce a set of images and then decompose these generated images into albedo, diffuse shading and glossy effects using an off-the-shelf, self-supervised network~\cite{forsyth2020intrinsic}.
We then search for style code edits by prompting StyleGAN to produce images that (a) are diverse, but (b) have the same albedo (and so geometry and material) as the original generated images.
Style codes are edited by adding constant vectors to the style code. This search procedure results in a robust set of directions that generalize to new images generated by the StyleGAN network.

Our search selects the most effective relighting directions in a data-driven manner. First, we use multiple intrinsic image methods to generate a large
pool of directions. Next, we prune this set of directions by measuring their impact on albedo, selecting only those that result in moderate albedo change while maintaining high diversity.
Finally, we apply a forward selection process that considers both low albedo change and high diversity to identify the best directions.

Although there are several distinct lightings or resurfacings of scenes in the training data, there has been no evidence thus far, that a StyleGAN network can produce multiple lightings
or resurfaces of scenes unprovoked. We believe that this work is the first demonstration of using style code edits (obtained without any supervision) to prompt the network and produce such variations to an image.
Our approach generates images with realistic lighting effects, including cast shadows, soft shadows, inter-reflections, and glossy effects. Importantly, we observe that style code edits produce consistent effects across images. For instance, as seen in Fig.~\ref{fig:teaser}, adding the first and second directions tends
to switch on bedside lamps (columns Relit-1 and Relit-2), while adding the fourth direction increases the light intensity from outside the window (column Relit-4). Since StyLitGAN can generate any image that a vanilla StyleGAN can, but also generate images that are out of distribution, one would expect FID scores to increase over StyleGAN, which we do observe, indicating that the distribution of images generated by StyLitGAN has a strict larger support. Finally, we provide a qualitative analysis of images generated by StyLitGAN.

 \section{Related Work}

{\noindent \bf Image manipulation:}  A significant literature deals with manipulating and editing
images~\cite{reinhard2001color, hertzmann2001image, efros2001image, liao2012subdivision, deshpande2017learning,
  zhu2017unpaired, gatys2016image, bhattad2020cut, ulyanov2018deep, park2019semantic}. Editing procedures for generative
image models~\cite{goodfellow2014generative} are important, because they demand compact image representations with 
useful, disentangled, interpretations. StyleGAN~\cite{karras2019style, Karras2019stylegan2, karras2021alias} is currently
de facto state-of-the-art for editing generated images, likely because its mapping of initial noise vectors to style codes which control entire feature layers produces
latent spaces that are heavily disentangled and so easy to manipulate. Recent editing methods include~\cite{voynov2020unsupervised, wu2021stylespace, shen2020interfacegan, zhu2020indomain, chong2021stylegan,
  richardson2021encoding}, with a survey in~\cite{xia2021survey}.  The architecture can be adapted to incorporate spatial priors for
authoring novel and edited images~\cite{ling2021editgan, tewari2020stylerig, epstein2022blobgan}.  In contrast to this
literature, we show how to fix one physically meaningful image factor while changing another.   Doing so is difficult because the latent spaces are not perfectly disentangled, and we must produce a diverse set of changes in the other factor.

{\noindent \bf Relighting using StyleGAN:} Relighting faces using StyleGAN can be achieved with Stylerig~\cite{tewari2020stylerig} but this method requires a 3D morphable face model. In contrast, StyLitGAN does not require a 3D model and can be extended to complex indoor scenes, which is not possible with Stylerig. Yang et al.~\cite{yang2019semantic} uses semantic label attributes to train a binary classifier to find latent space directions that represent indoor and natural lighting, but this method cannot produce diverse relighting effects. In contrast, StyLitGAN generates a wide range of diverse and realistic relighting effects without requiring any labeled attributes.

 StyleFlow~\cite{abdal2021styleflow} and GAN control~\cite{shoshan2021gan}, require a parametric model to express lighting, such as spherical harmonics. These methods are limited to relighting faces and cannot be applied to rooms. In contrast, StyLitGAN can produce images of relit or recolored rooms. Note: rooms are more challenging to relight than faces due to significant long-scale inter-reflection effects, diverse shadow patterns, stylized luminaires, stylized surface albedos, and surface brightnesses that are not a function of surface normal alone. These factors make it difficult to apply environment maps or spherical harmonic directly to rooms. Additionally, none of these methods have the ability to resurface or recolor rooms.

\begin{figure*}[t!]
    \centering
    \includegraphics[width=0.98\linewidth]{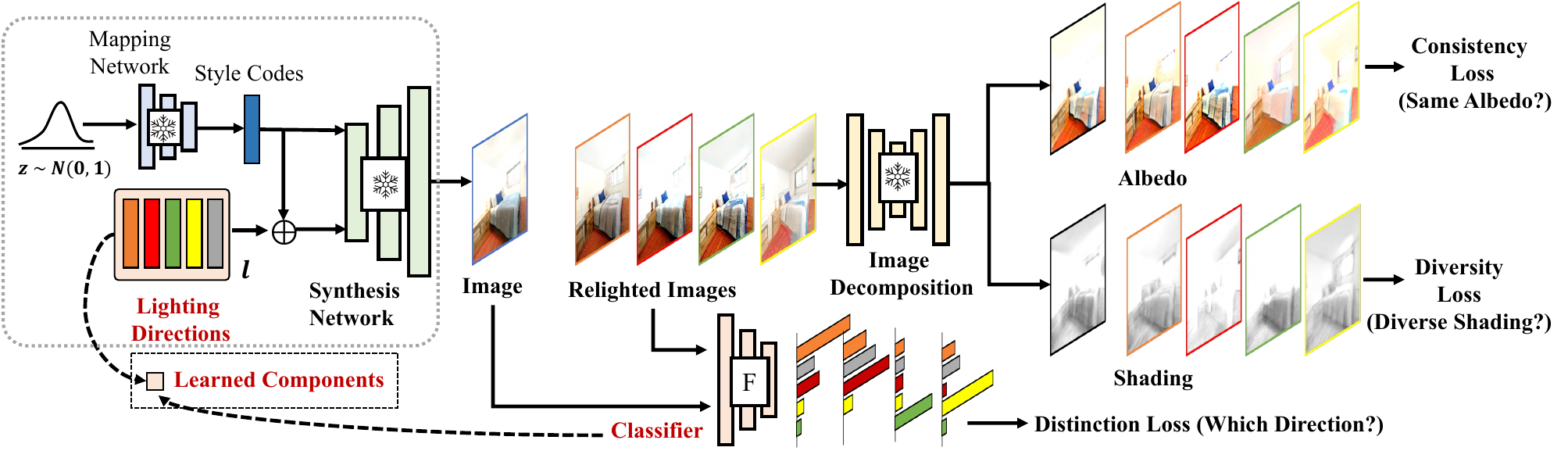}
    \vspace{-5pt}
    \caption{How StyLitGAN works: We generate an image from a random Gaussian noise using a pretrained StyleGAN. We also generate novel relighted images (16 in our case) of the same using randomly initialized latent directions ($\vect{d}$) that are added to $\vect{w}^+$ latent style codes. We train a classifier (F) that takes in all the pairs of relighted and original images and predicts the relighting direction applied to them. We apply a distinction loss and jointly update the latent directions and the classifier.  Next, we generate the decomposition of these images from a pretrained decomposition model (D). We then apply losses that force StyLitGAN to find latent directions so that the albedo does not change (consistency loss), but the image does (diversity loss). %
    }
    \label{fig:approach}
      \vspace{-15pt}
  \end{figure*}

{\noindent \bf Other Face Relighting} methods use	carefully collected supervisory data from light-stages or parametric spherical harmonics~\cite{sun2019single,   zhou2019deep, nestmeyerlearning, sengupta2021light,	
  pandey2021total}. ShadeGAN~\cite{pan2021shadegan}, Rendering with Style~\cite{chandran2021rendering} and Volux-GAN~\cite{tan2022volux} uses a volumetric rendering approach to learn the 3D structure of the face and the illumination encoding. Volux-GAN~\cite{tan2022volux} also requires image	
decomposition from ~\cite{pandey2021total} that is trained using carefully curated light-stage data. In comparison, we neither require any explicit 3D modeling of the scene nor labeled and curated data for training the image decomposition model.

{\noindent \bf Image Decomposition:} Current image decomposition methods~\cite{li2018cgintrinsics, bi20151, janner2017self, 
  fan2018revisiting, forsyth2020intrinsic}. Each scores very well in current evaluations of WHDR (standard metric for intrinsic images due to~\cite{bell2014intrinsic}; recent summary of SOTA performance in~\cite{forsyth2020intrinsic})

\section{Approach}
  \label{directions}
We follow convention and manipulate StyleGAN~\cite{Karras2019stylegan2} by adjusting the $\vect{w^+}$ latent variables.  We do not modify StyleGAN, but instead, seek a set of
lighting directions $\vect{d}_i$ which are independent of $\vect{w^+}$ and which have desired effects on the generated image.
We obtain these directions by constructing losses that capture the desired outcomes, then search for directions
that minimize these losses.  We find all directions at once and use 2000 randomly generated images for this search. Once found, these lighting directions are applicable to all other generated images.  Our search
procedure only
sees each image once. 
Fig.~\ref{fig:approach} summarizes our procedure and we call our model StyLitGAN. 

\noindent{\bf Base StyleGAN Models:} We use baseline pretrained models from ~\cite{yu2021dual} that use a dual-contrastive loss to train StyleGAN for bedrooms, faces and churches. We also use baseline pretrained StyleGAN2 models from ~\cite{epstein2022blobgan} for conference rooms and kitchen, dining and living rooms. 

\noindent{\bf Decomposition:}   
We decompose images into albedo, shading and gloss maps (gloss; only when available) as $A \times S + G$, where $A$ models
albedo effects and $S$ and $G$ model shading and gloss effects respectively.
We use the method of Forsyth and Rock~\cite{forsyth2020intrinsic}, which is easily adapted because it uses only samples from statistical
models that are derived from Land's Retinex theory~\cite{land1977retinex} and is self-supervised.  We can construct many decompositions using their approach by changing the statistical spatial model parameters. We evaluate several such decomposition models under several hyperparameters settings and create a large pool of relighting directions. We finalize our directions using a forward selection process that provides minimal albedo and geometry shift with a large relighting diversity (Section~\ref{sec:model_selection}).

\noindent\noindent{\bf Relighting} a scene should produce a new, realistic, image where the shading has changed but the albedo has not. 
Write $I(\vect{w^+})$ for the image produced by StyleGAN given style codes $\vect{w^+}$, and $A(I)$, $S(I)$ and $G(I)$) for the
albedo, shading and gloss respectively recovered from image $I$.
We search for multiple directions $\vect{d}_i$ such that: (a) $A(I(\vect{w^+}+ \vect{d}_i))$ is very close to
$A(I(\vect{w^+}))$ -- so the image is a relighted version of $I(\vect{w^+})$, a property we call {\bf persistent consistency};
(b) the images produced by the different directions are linearly independent -- {\bf relighting diversity};
(c) which direction was used can be determined from the image, so that different directions have visibly distinct
effects  -- {\bf distinctive relighting}; and
(d) the new shading field is not strongly correlated to the albedo  -- {\bf independent relighting}.
Not every shading field can be paired with a given albedo, otherwise, there would be nothing to do.
We operate on the assumption that edited $\vect{w^+}$ will result in realistic
images~\cite{chong2021jojogan}.   

\noindent{\bf Recoloring:}  Alternatively, we may wish to edit scenes where the colors of materials of objects have changed, but the lighting hasn't. Because shading conveys a great deal of information about shape, we can find these edits using modified losses by seeking consistency in the shading field.
\vspace{-12pt}
\paragraph{Persistent Consistency:} The albedo decomposition of both the relighted scene: $A_R=A(I(\vect{w^+}+\vect{d}_i))$ and the original: $A_O=A(I(\vect{w^+}))$ must be the same; where R refers to relighted images and O refers to StyleGAN generated images. We use a Huber loss and a perceptual feature loss ~\cite{johnson2016perceptual, zhang2018perceptual} from a VGG feature extractor ($\Phi$) ~\cite{simonyan2014very}  at various feature layers ($j$) to preserve persistent effects (geometry, appearance and texture) in the scene. %
\begin{equation}
    \small
    \mathcal{L}_{const}(A_O,A_R) = \left\{ \begin{array}{cl}
\frac{1}{2} \left[A_O-A_R\right]^2 & \hspace{-15pt}\text{for\  }|A_O-A_R| \le \delta, \\
\delta \left(|A_O-A_R|-\delta/2\right) & \text{otherwise.}
\end{array}\right.
\end{equation}
\begin{equation}
\label{eq:percep}
    \mathcal{L}_{per}(A_O,A_R) = ||\Phi_{j}(A_O) - \Phi_j(A_R) ||_2.
    \vspace{-10pt}
\end{equation}
\vspace{-10pt}
\paragraph{Relighting Diversity:} We want the set of relighted images produced by the directions
to be diverse on a long scale so that regions that were in shadow in one image might be bright in another.
For each $S(\vect{w}^+ +\vect{d}_i)$, we stack the two shading and gloss: $S$ and $G$ and compute a smoothed and downsampled vector
$\vect{t}_i$ from these maps.   We then compute $\mathcal{L}_{div}(S, G)$ (diversity loss)  which compels these $\vect{t}_i$ to be linearly independent and encourages diversity in relighting.
\begin{equation}
\label{eq:diversity}
\small
    \begin{array}{cc}
        \mathcal{L}_{div}(S, G) = -\log\det N  
    \end{array} 
  \end{equation}
  
  {where\ $i^{th}$ \& $j^{th}$ component of} N \text{\ is\ } 
  $t_i^\intercal t_j$
  \vspace{-10pt}
\paragraph{Distinctive Relighting:} A network might try to cheat by making minimal changes to the image.  Directions
$\vect{d}_i$ should have the property that $\vect{d}_i$ is easy to impute from $I(\vect{w^+}+\vect{d}_i)$. We train
a classifier joint with the search for directions.  This classifier accepts $I(\vect{w^+})$ and $I(\vect{w^+}+\vect{d}_i)$
and must predict $i$.  The cross-entropy of this classifier supplies our loss:
\begin{equation}
\label{eq:distinction}
\small
\centering
\begin{aligned}
        \min_{l, F}{\mathcal{L}_{dist}}(I(\vect{w^+}),\  I(\vect{w^+}+d_i)) \\ =  -\sum_{i=1}^Md_{i}\log F(I(\vect{w^+}),\  I(\vect{w^+}+d_i))
\end{aligned}
\end{equation}
\vspace{-10pt}
\paragraph{Saturation Penalty:} Our diversity loss might cheat and obtain high diversity by
generating blocks of over-saturated or under-saturated pixels. To discourage these effects, we apply a saturation penalty over number of pixels within a certain threshold.  
\begin{equation}
\begin{aligned}
   \mathcal{L}_{sat} = \lambda_{oversat}  \big[ \frac{1}{H\times W}\sum_{i=1}^{H} \sum_{j=1}^{W} max(0, I_{i,j} - s)^2 \big] \\ 
+ \lambda_{undersat} \big[ \frac{1}{H\times W} \sum_{i=1}^{H} \sum_{j=1}^{W} max(0, s - I_{i,j})^2 \big]
\end{aligned}
\end{equation} 
where $\lambda_{oversat}$ and $\lambda_{undersat}$ are the penalty weight for over-saturation and under-saturation respectively, $H$ and $W$ are the height and width of the images, $I_{i,j}$ is the pixel intensity at pixel location $(i,j)$, and $s$ is the saturation threshold (i.e., the maximum allowed pixel intensity). The penalty is computed as the mean squared difference between the pixel intensity and the saturation threshold.

{\bf Recoloring} requires swapping albedo and shading components in all losses, except we do not use decorrelation loss while recoloring.
Obtaining good results requires quite a careful choice of loss weights ($\lambda$ coefficients). We experiment with several $\lambda$ coefficients for both these edits (Section~\ref{sec:model_selection} and supplementary).

\section{Model and Directions Selection}
\label{sec:model_selection}
\begin{figure}[t!]
\centering
    \includegraphics[width=0.8\linewidth]{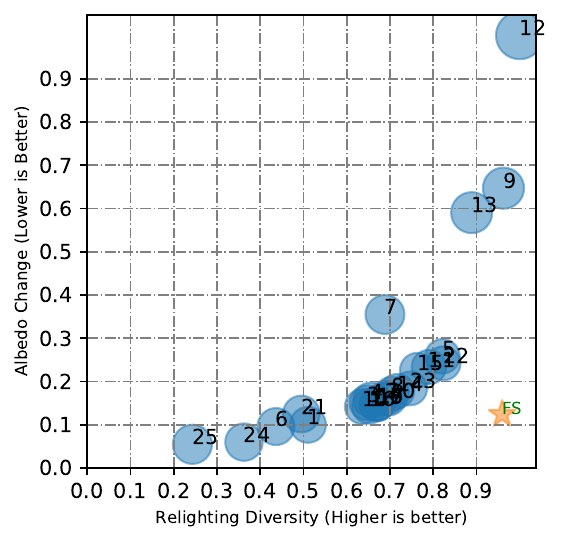}
    \vspace{-5pt}
  \caption{Each model produces 16 directions.  Models differ by choice of hyperparameters and intrinsic image decomposition.  We evaluate models by albedo change and by diversity,
averaged across a small fixed validation set of test scenes.  As the figure shows, there is typically
a payoff, but some models are not admissible.  Figure~\ref{fig:model_ablation} shows examples from some
of the models considered here.  We exclude inadmissible models, then
pool all directions from all other models, and apply a forward selection procedure (section~\ref{sec:model_selection}).
This yields 16 strong relighting directions (the star).}
    \label{fig:model_selection}
\vspace{-10pt}
  \end{figure}

We prompt StyleGAN to to find style code directions that:
(a) do not change albedo; and (b) strongly change the image. We use a variety of different image decomposition models to obtain directions across multiple different hyperparameter settings. We have no particular reason to believe that a single model will give only good directions, or all good directions.  We then find a subset of admissible models.  We must choose admissible models using a plot of albedo change versus diversity because there is no way to
weigh these effects against one another. However, relatively few methods are admissible -- see Figure~\ref{fig:model_selection}.  We then pool all directions from all of the admissible models, and use forward selection to find a small set (16 in this work) of polished directions in this pool. 
\vspace{-10pt}
\paragraph{Scoring albedo change:}  We use SuperPoints~\cite{hui2021superpoint} to find 100 interest points in
the original StyleGAN-generated image.  Around each interest point, we form a
$8 \times 8$ patch.  We then compare these patches with patches in the same locations for multiple different relightings of that image.
If the albedo in the image does not change, then each patch will
have the same albedo but different lighting. 

Given two color image patches $\vect{p}$ and $\vect{q}$, viewed under different lights, we must measure the difference between their albedos $d_a(\vect{p}, \vect{q})$. Write $p_{ij}$
for the RGB vector at the $i$, $j$'th location ($1 \leq i\leq M$, $1 \leq j \leq N$) and write
$p_{ij,k}$ for the $k$’th RGB component at that location.  The intensity of the light
may change without the albedo changing, so  this problem is
homogeneous (i.e. for $\lambda, \mu > 0$,
$d_a(\vect{p}, \vect{q}) = d_a(\lambda \vect{p}, \mu \vect{q})$).
This suggests using a cosine distance. We assume that the illumination intensity changes, but not the illumination
color. The illumination fields may vary across each patch, but the  patches are small. This allows the illumination
field to be modeled as a linear function, so that there are albedos $\vect{a}$, $\vect{b}$
such that $p_{ij} = (p_x i + p_y j + p_c) a_{ij}$  
and $q_{ij} = (q_x i + q_y j + q_c)b_{ij}$. In turn, if the two patches have similar albedo,
there will be $p_x$ etc. such that $p_{ij}'(q_x, q_y, q_c) = (q_x i+q_y j +q_c)p_{ij}$ is the same as
$q_{ij}' (p_x, p_y, p_c) = (p_x i + p_y j + p_c)q_{ij}$ . We measure the cosine distance
\[
d_a(\vect{p}, \vect{q})=1-\max_{p_x, \ldots q_c}\frac{\sum_{ijk} p_{ijk}' q_{ijk}'}{\sqrt{\sum_{ijk}
    (p_{ijk}')^2}\sqrt{\sum_{ijk} (p_{ijk}')^2}}
\]
The relevant maximum can be calculated by analogy with canonical correlation analysis (Supplementary).

\vspace{-10pt}
\paragraph{\bf Scoring Lighting Diversity:} 
Illumination cone theory~\cite{belhumeur1998set} yields that any non-negative linear combination of $k$ shadings is a physically plausible shading. To determine if an image is new, we relax the non-negativity constraint and so must ensure that it cannot be expressed as a linear combination of existing images. In turn, we seek a measure of the linear independence
of a set of images.  This measure should: be large when there is a strong linear
dependency; and not grow too fast when the images are scaled.
Write $\vect{x}_i$ for the $i$'th image, and ${\cal X}$ for the matrix whose $i$, $j$'th component is $\vect{x}_i\vect{x}_j$.  Then
$-\log \det {\cal X}$ is very large when the $\vect{x}_i$ is close to linearly
dependent, but do not scale too fast when the images are scaled.

\vspace{-10pt}
\paragraph{\bf Decomposition Models Investigated:}
We searched 25 instances in total obtained with different hyperparameter settings from three families of decomposition.
The first family is the SOTA unsupervised model of~\cite{forsyth2020intrinsic}, which decomposes images
into albedo and shading using example images drawn from statistical models.  The second is a variant of
that family that decomposes into albedo, shading and gloss decomposition.
The third is an albedo, shading and gloss decomposition that models fine edges in the albedo rather than the shading
field.   These models were chosen to represent a range of possible decompositions, but others could yield better
results. The key point is that we can choose a model from a collection by a rational process.

\begin{figure}[t!]
\scriptsize
  \centering
  \footnotesize
  \setlength\tabcolsep{0.2pt}
  \renewcommand{\arraystretch}{0.1}

  \begin{tabular}{cccccc}
\multicolumn{1}{c}{\rotatebox{90}{\hspace{5pt} Model 25}}  &     \includegraphics[width=0.18\linewidth]{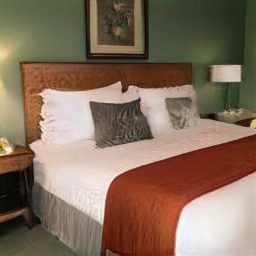} & \includegraphics[width=0.18\linewidth]{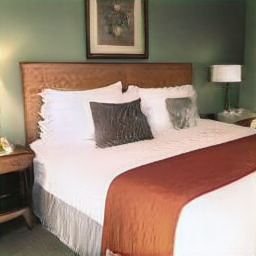} & \includegraphics[width=0.18\linewidth]{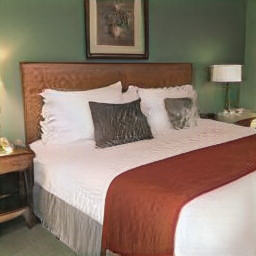} & \includegraphics[width=0.18\linewidth]{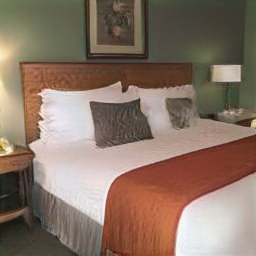} &  
 \includegraphics[width=0.18\linewidth]{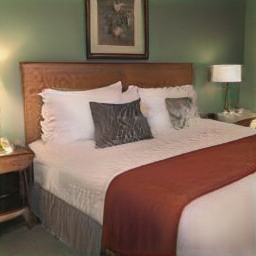} 
    \\
\centering{\rotatebox{90}{\hspace{5pt} Model 1}} & \includegraphics[width=0.18\linewidth]{figures/model_ablation/FS_orig.png} &  \includegraphics[width=0.18\linewidth]{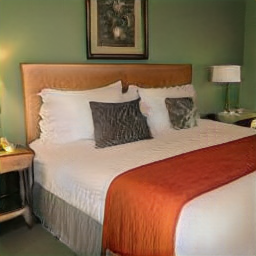} & \includegraphics[width=0.18\linewidth]{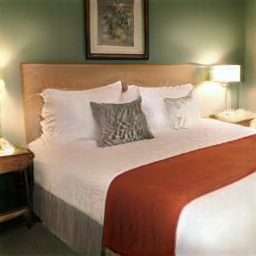} & \includegraphics[width=0.18\linewidth]{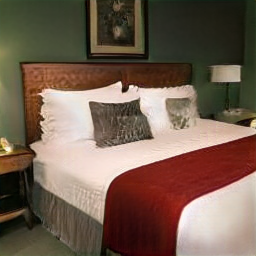} &  
 \includegraphics[width=0.18\linewidth]{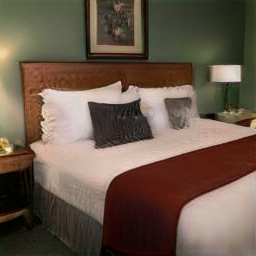} 
    \\
\centering\rotatebox{90}{\hspace{5pt} Model 9} &     \includegraphics[width=0.18\linewidth]{figures/model_ablation/FS_orig.png} & \includegraphics[width=0.18\linewidth]{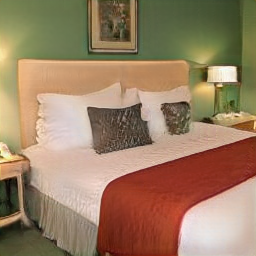} & \includegraphics[width=0.18\linewidth]{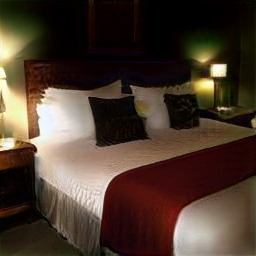} & \includegraphics[width=0.18\linewidth]{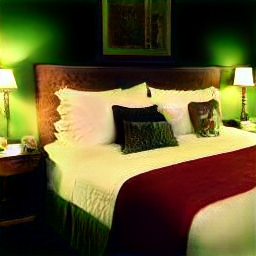} &  
 \includegraphics[width=0.18\linewidth]{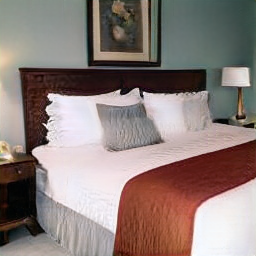} 
    \\
\centering\rotatebox{90}{\hspace{5pt} FS (final)\hfil} &     \includegraphics[width=0.18\linewidth]{figures/model_ablation/FS_orig.png} &  \includegraphics[width=0.18\linewidth]{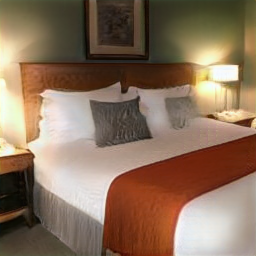} & \includegraphics[width=0.18\linewidth]{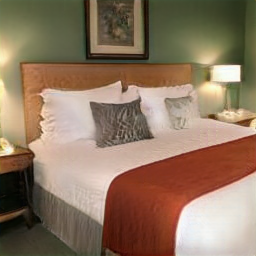} & \includegraphics[width=0.18\linewidth]{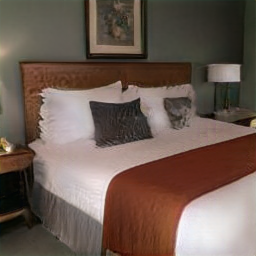} &  
 \includegraphics[width=0.18\linewidth]{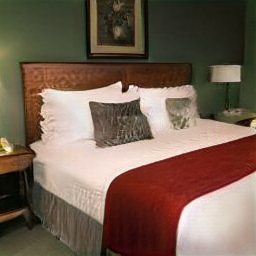} 
    \\
   &  Image & Relit - 1  &  Relit - 2  & Relit - 3   & Relit - 4 
        \end{tabular}
      \caption{The {\bf bottom} row shows scene relightings obtained using our final,  forward selected, set
of directions (star of Figure~\ref{fig:model_selection}).  For comparison, we show scene relightings from each of the
models of that figure (model numbers correspond to numbers on that figure).  Note how most
models are capable of producing some good directions, but not all directions from a given model
are good.}
\vspace{-10pt}
\label{fig:model_ablation}        
    \end{figure}

\vspace{-10pt}
\paragraph{\bf Selecting Directions:} Our approach for selecting directions involves creating a scatter plot of 25 instances with various hyperparameters and image decomposition models. We find 16 directions for each instance in our final experiments and the search for 16 directions takes about 14 minutes on an A40 GPU.  We experimented with different numbers of directions of order $2^n$ for n={2, 3, 4, 5, 6, 7} and found that 16 directions (at n=4) strike a better balance between relighting diversity and albedo change. However, finding multiple directions is challenging because the search space is complex and high-dimensional, and we lack ground truth to supervise the search. Therefore, we apply a two-step process to find effective directions and filter out any bad directions.

We first identify and discard inadmissible models that are located behind the Pareto frontier. We then select the top 10 admissible models based on their average albedo change when applied over a large set of fixed validation images. Our goal is to select the best relighting directions from these admissible models. To achieve this, a forward selection process is employed, which involves selecting a subset of directions from the set of admissible directions.
\vspace{-10pt}
\paragraph{\bf Forward Selection Process:} To select the best directions, we begin by selecting all directions from the admissible models, resulting in 160 directions from 10 models. These directions are then filtered to remove ``bad" directions that produce relighting similar to the original image or shading that does not vary across pixels, resulting in 108 directions.

Next, we use a greedy process to select the best 16 directions from the remaining 108 directions. We evaluate each direction one at a time and add it to the pool if it provides a large diversity score while incurring a small penalty for large albedo change. This process continues until the desired number of directions are selected. The forward selection process is fast and efficient, taking less than a minute.

The resulting scores from the forward select 16 directions are marked with a \emph{star} in golden color in Fig.~\ref{fig:model_selection}. The directions obtained with this process are significantly better than individual models alone. A qualitative ablation is in Fig.~\ref{fig:model_ablation}.

\begin{figure*}[htpb!]
  \footnotesize
  \setlength\tabcolsep{0.1pt}
  \renewcommand{\arraystretch}{0.1}
  \begin{tabular}{cccccccc}
        Generated Image & Relit - 1 ($+ d_1$) &  Relit - 2  ($ + d_2$) & Relit - 3  ($+ d_3$) & Relit - 4   ($ + d_4$)& Relit - 5  ($ + d_5$) & Relit - 6  ($ + d_6$) & Relit - 7  ($ + d_7$) \\
        \includegraphics[width=0.125\linewidth]{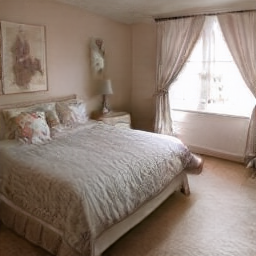} &
    \includegraphics[width=0.125\linewidth]{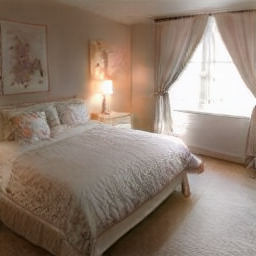} &
    \includegraphics[width=0.125\linewidth]{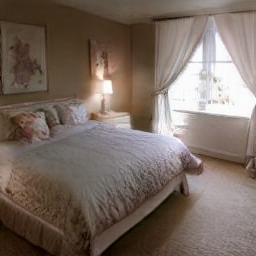} &
    \includegraphics[width=0.125\linewidth]{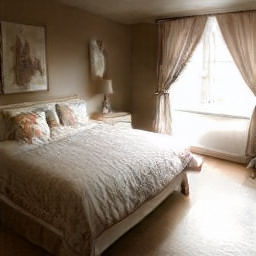} &
    \includegraphics[width=0.125\linewidth]{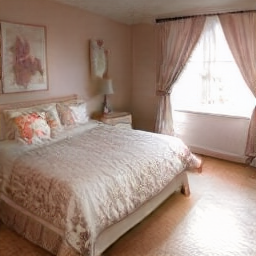} &
    \includegraphics[width=0.125\linewidth]{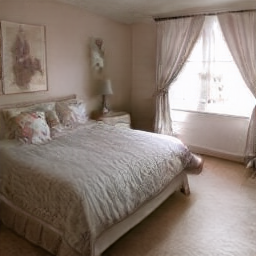} & 
    \includegraphics[width=0.125\linewidth]{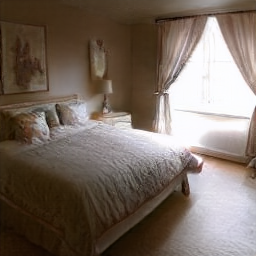} &
    \includegraphics[width=0.125\linewidth]{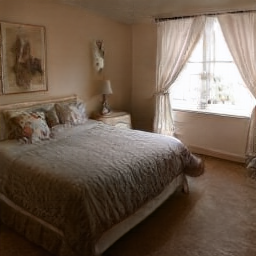} 
        \\
    \includegraphics[width=0.125\linewidth]{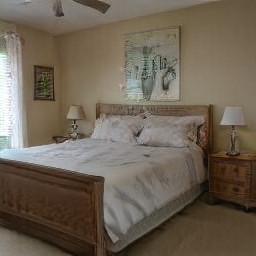} &
    \includegraphics[width=0.125\linewidth]{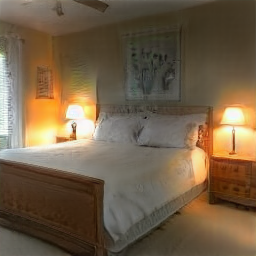} &
    \includegraphics[width=0.125\linewidth]{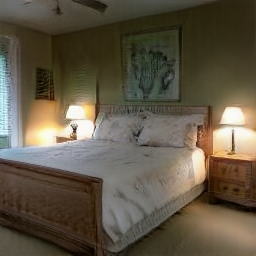} &
    \includegraphics[width=0.125\linewidth]{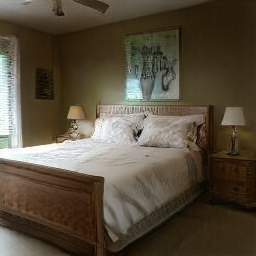} &
    \includegraphics[width=0.125\linewidth]{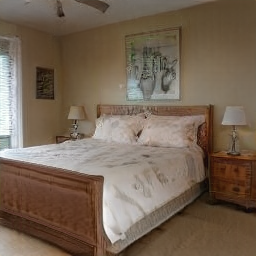} &
    \includegraphics[width=0.125\linewidth]{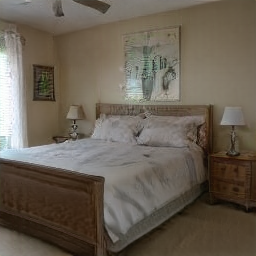} &
    \includegraphics[width=0.125\linewidth]{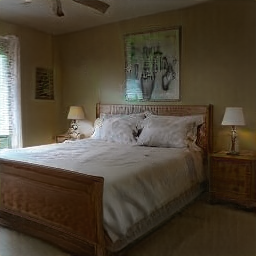} &
    \includegraphics[width=0.125\linewidth]{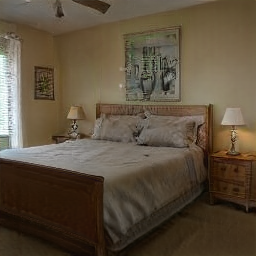}
        \\
    \includegraphics[width=0.125\linewidth]{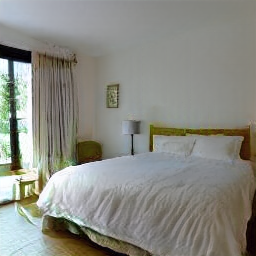} &
    \includegraphics[width=0.125\linewidth]{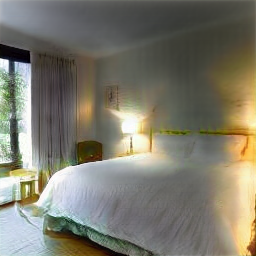} &
    \includegraphics[width=0.125\linewidth]{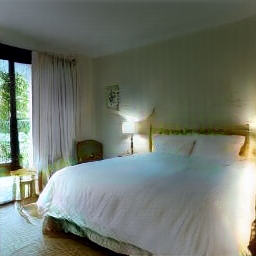} &
    \includegraphics[width=0.125\linewidth]{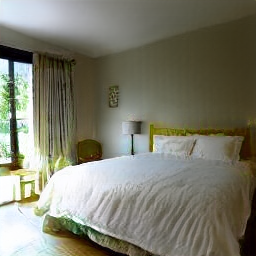} &
    \includegraphics[width=0.125\linewidth]{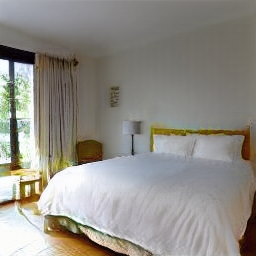} &
    \includegraphics[width=0.125\linewidth]{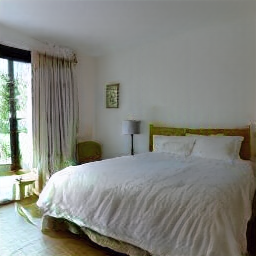} &
    \includegraphics[width=0.125\linewidth]{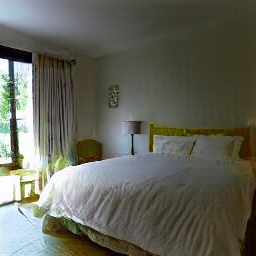} &
    \includegraphics[width=0.125\linewidth]{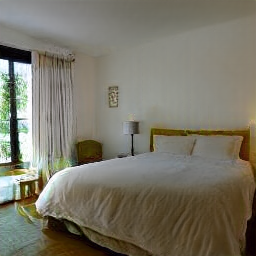}
            \\
    \includegraphics[width=0.125\linewidth]{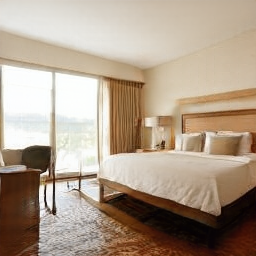} &
    \includegraphics[width=0.125\linewidth]{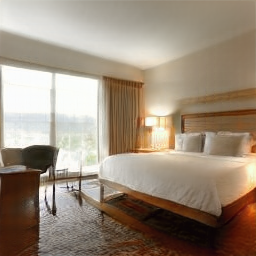} &
    \includegraphics[width=0.125\linewidth]{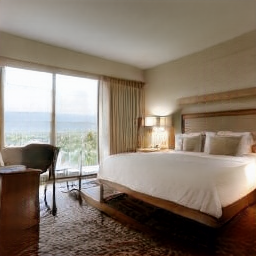} &
    \includegraphics[width=0.125\linewidth]{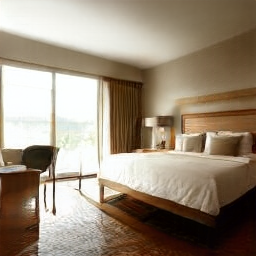} &
    \includegraphics[width=0.125\linewidth]{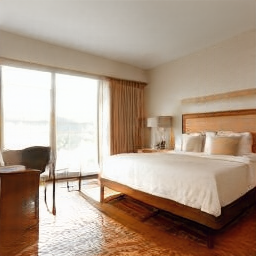} &
    \includegraphics[width=0.125\linewidth]{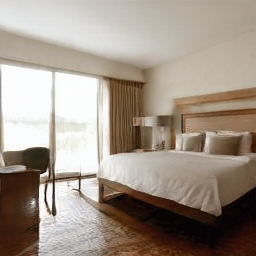} &
    \includegraphics[width=0.125\linewidth]{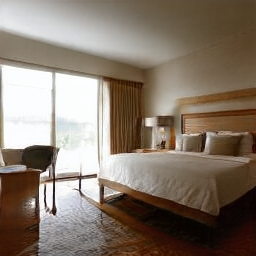} &
    \includegraphics[width=0.125\linewidth]{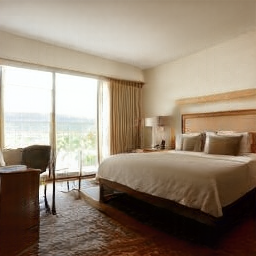}
    \\
    \includegraphics[width=0.125\linewidth]{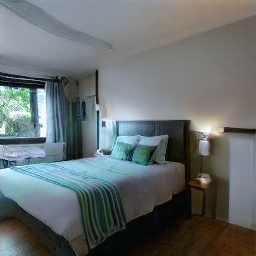} &
    \includegraphics[width=0.125\linewidth]{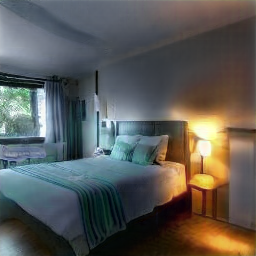} &
    \includegraphics[width=0.125\linewidth]{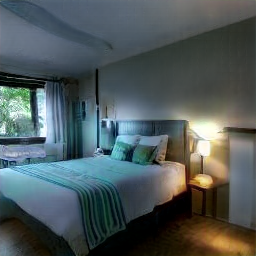} &
    \includegraphics[width=0.125\linewidth]{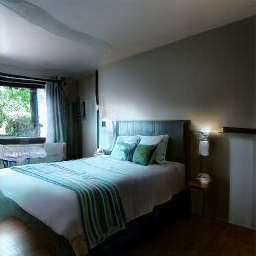} &
    \includegraphics[width=0.125\linewidth]{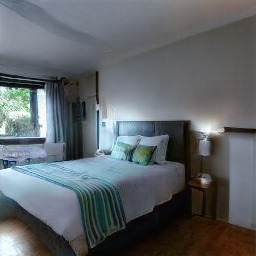} &
    \includegraphics[width=0.125\linewidth]{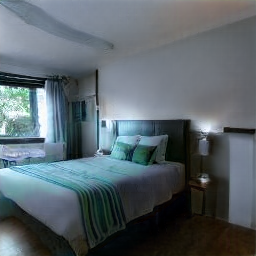} &
    \includegraphics[width=0.125\linewidth]{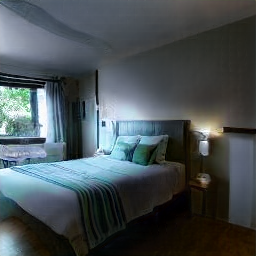} &
    \includegraphics[width=0.125\linewidth]{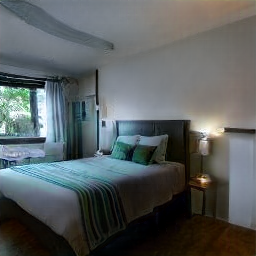}
        \\
        \includegraphics[width=0.125\linewidth]{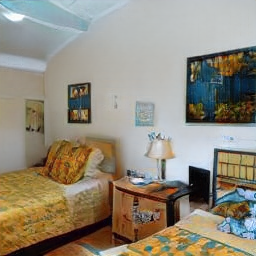} &
    \includegraphics[width=0.125\linewidth]{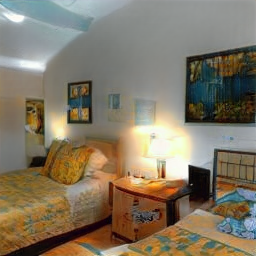} &
    \includegraphics[width=0.125\linewidth]{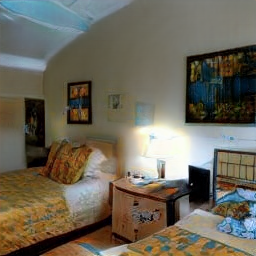} &
    \includegraphics[width=0.125\linewidth]{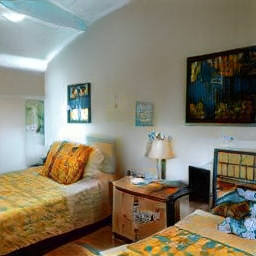} &
    \includegraphics[width=0.125\linewidth]{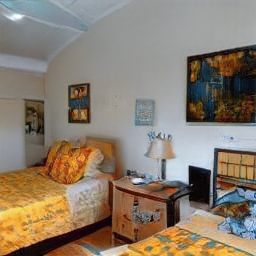} &
    \includegraphics[width=0.125\linewidth]{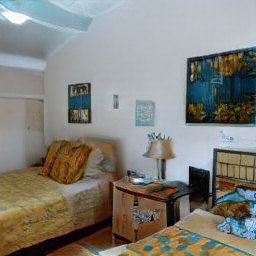} &
    \includegraphics[width=0.125\linewidth]{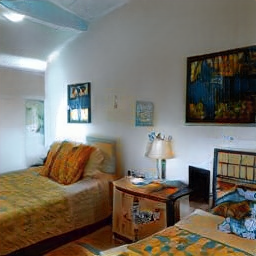} &
    \includegraphics[width=0.125\linewidth]{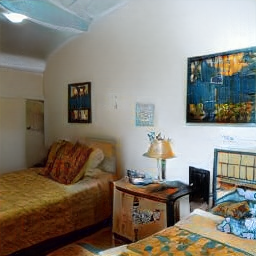}
        \\
        \includegraphics[width=0.125\linewidth]{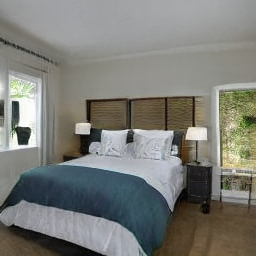} &
    \includegraphics[width=0.125\linewidth]{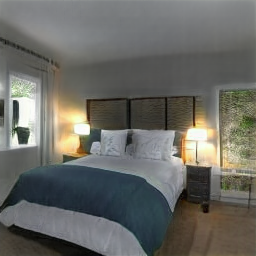} &
    \includegraphics[width=0.125\linewidth]{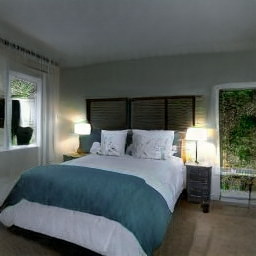} &
    \includegraphics[width=0.125\linewidth]{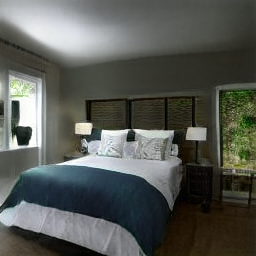} &
    \includegraphics[width=0.125\linewidth]{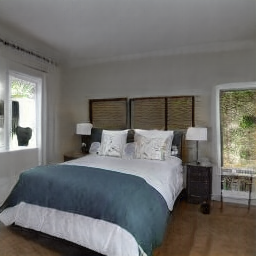} &
    \includegraphics[width=0.125\linewidth]{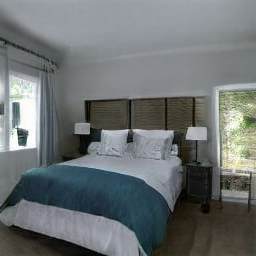} &
    \includegraphics[width=0.125\linewidth]{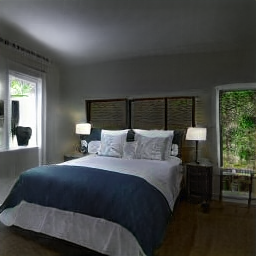} &
    \includegraphics[width=0.125\linewidth]{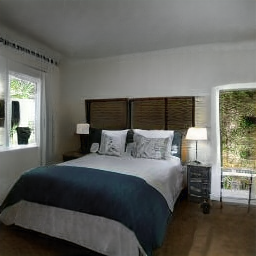}
        \\
        \includegraphics[width=0.125\linewidth]{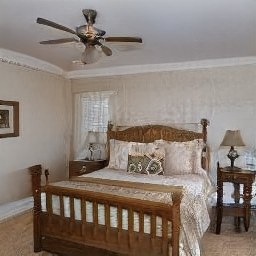} &
    \includegraphics[width=0.125\linewidth]{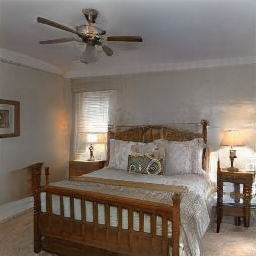} &
    \includegraphics[width=0.125\linewidth]{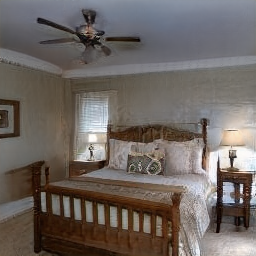} &
    \includegraphics[width=0.125\linewidth]{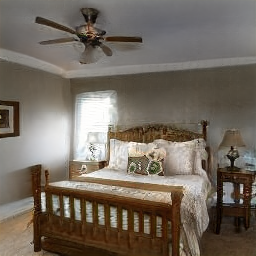} &
    \includegraphics[width=0.125\linewidth]{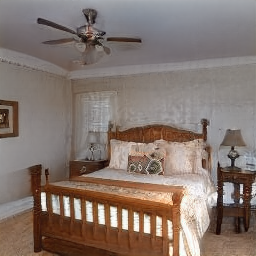} &
    \includegraphics[width=0.125\linewidth]{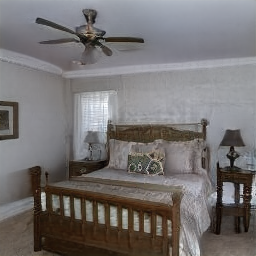} &
    \includegraphics[width=0.125\linewidth]{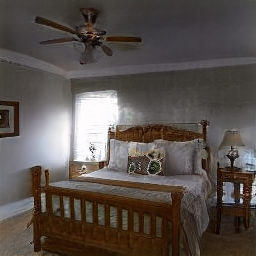} &
    \includegraphics[width=0.125\linewidth]{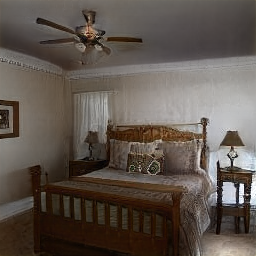}
        \\
        \includegraphics[width=0.125\linewidth]{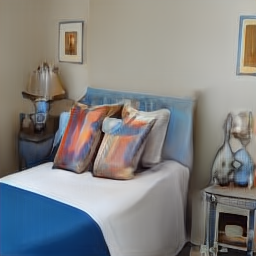} &
    \includegraphics[width=0.125\linewidth]{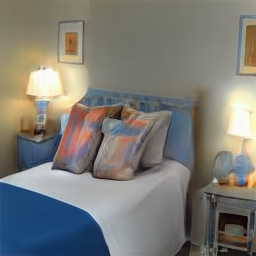} &
    \includegraphics[width=0.125\linewidth]{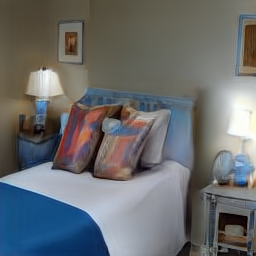} &
    \includegraphics[width=0.125\linewidth]{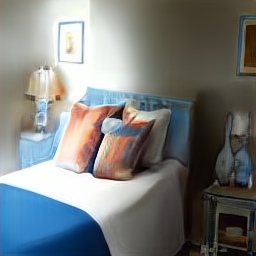} &
    \includegraphics[width=0.125\linewidth]{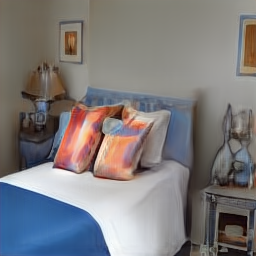} &
    \includegraphics[width=0.125\linewidth]{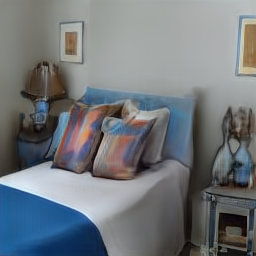} &
    \includegraphics[width=0.125\linewidth]{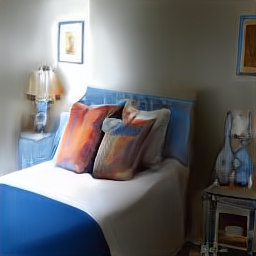} &
    \includegraphics[width=0.125\linewidth]{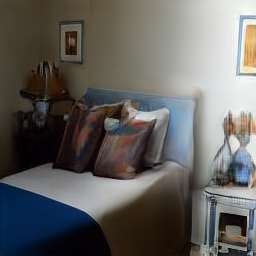}
  
        \end{tabular}
  \caption{{\bf First columns} images generated by the original StyleGAN.  {\bf Other columns} show
    images obtained from $\vect{w^+}+\vect{d}_i$ for our relighting directions added to the style codes ($\vect{w^+}$) corresponding to the
    the image on the first column's style codes ($\vect{w^+}$).  
    These directions have been chosen to fix albedo, but change shading.  Note:  each column appears to show the same scene that is the row with
    different illumination; lighting varies aggressively, and the individual latent direction $\vect{d}_i$ have persistent semantics -- each column corresponds to a type of illumination. Notice relighting changes around ceilings (when visible), walls, and bedside lamps. Another observation that can be made is that all relightings are with respect to world coordinates and not with respect to the camera coordinates.  
  \label{relighting}}
\end{figure*}

\begin{figure*}[t!]
  \footnotesize
  \setlength\tabcolsep{0.1pt}
  \renewcommand{\arraystretch}{0.1}
  \begin{tabular}{cccccccc}
        Generated Image & Resurf.- 1 ($+ d_1$) &  Resurf.- 2  ($ + d_2$) & Resurf.- 3  ($+ d_3$) & Resurf.- 4   ($ + d_4$)& Resurf.- 5  ($ + d_5$) & Resurf.- 6  ($ + d_6$) & Resurf.- 7  ($ + d_7$) \\
    \includegraphics[width=0.125\linewidth]{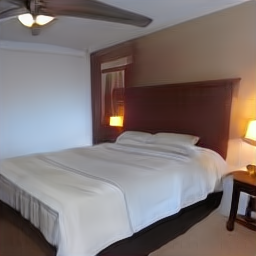} &
    \includegraphics[width=0.125\linewidth]{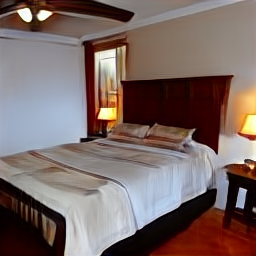} &
    \includegraphics[width=0.125\linewidth]{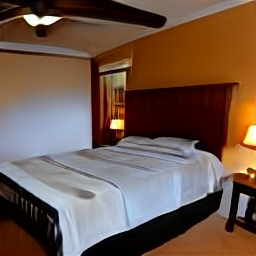} &
    \includegraphics[width=0.125\linewidth]{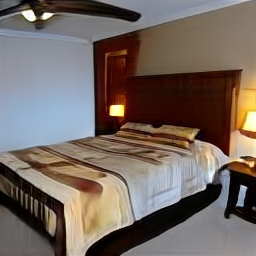} &
    \includegraphics[width=0.125\linewidth]{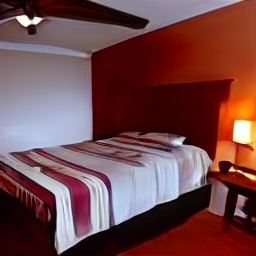} &
    \includegraphics[width=0.125\linewidth]{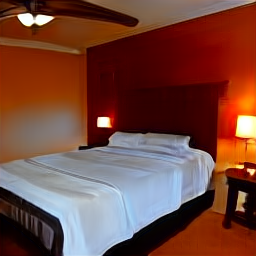} & 
    \includegraphics[width=0.125\linewidth]{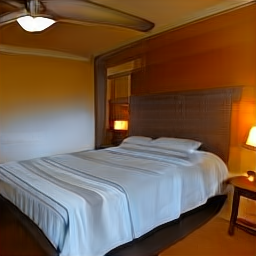} &
    \includegraphics[width=0.125\linewidth]{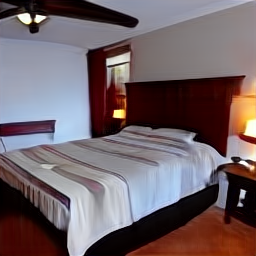} 
        \\
    \includegraphics[width=0.125\linewidth]{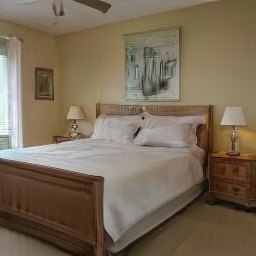} &
    \includegraphics[width=0.125\linewidth]{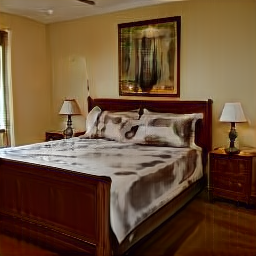} &
    \includegraphics[width=0.125\linewidth]{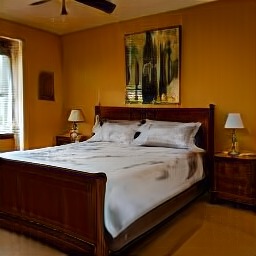} &
    \includegraphics[width=0.125\linewidth]{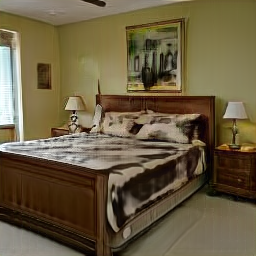} &
    \includegraphics[width=0.125\linewidth]{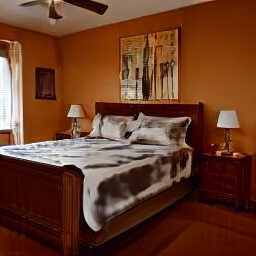} &
    \includegraphics[width=0.125\linewidth]{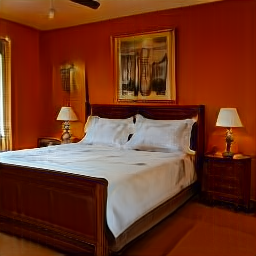} & 
    \includegraphics[width=0.125\linewidth]{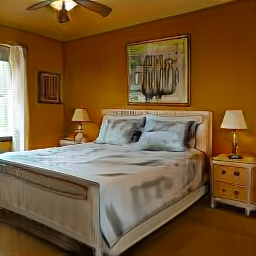} &
    \includegraphics[width=0.125\linewidth]{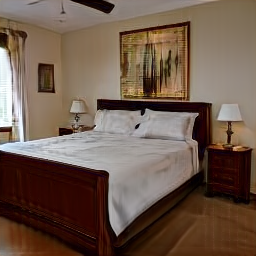} 
        \end{tabular}
\vspace{2pt}
 \caption{Resurfacing Generated Images. Instead of relighting images, we can generate resurfaced images by swapping our consistency and diversity loss. We apply diversity loss to change the albedo and consistency loss to maintain the shading and overall illumination. The {\bf first column} shows images generated by the original StyleGAN, and the {\bf other columns} show images obtained from $\vect{w^+}+\vect{d}_i$ for our resurfacing or recoloring directions. Each column shows the same scene as in the first column, but with varying surface colors and materials, while the individual latent direction $\vect{d}_i$ retains its semantics.
  \label{resurfacing}}
\end{figure*}

\begin{figure*}[t!]
  \footnotesize
  \setlength\tabcolsep{0pt}
  \renewcommand{\arraystretch}{0.1}
  \begin{tabular}{cccccccc}
    \includegraphics[width=0.125\linewidth]{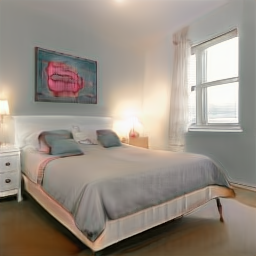} &
    \includegraphics[width=0.125\linewidth]{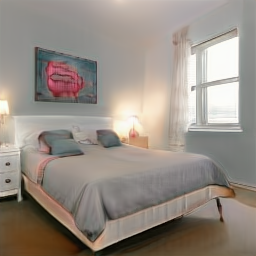} &
    \includegraphics[width=0.125\linewidth]{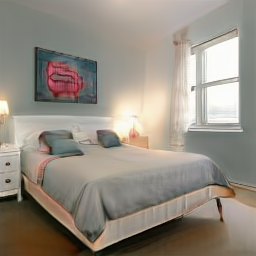} &
    \includegraphics[width=0.125\linewidth]{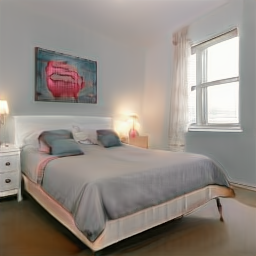} &
    \includegraphics[width=0.125\linewidth]{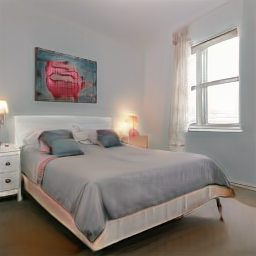} &
    \includegraphics[width=0.125\linewidth]{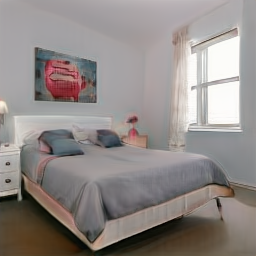} &
    \includegraphics[width=0.125\linewidth]{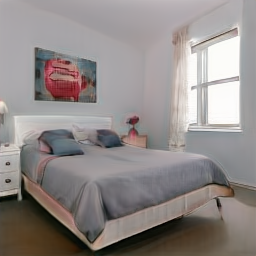} &
    \includegraphics[width=0.125\linewidth]{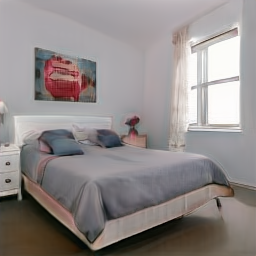} 
        \\
    \includegraphics[width=0.125\linewidth]{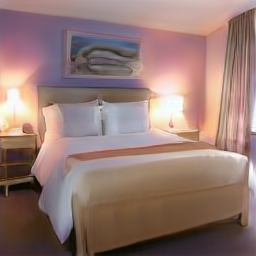} &
    \includegraphics[width=0.125\linewidth]{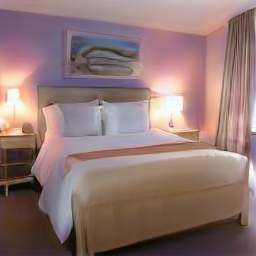} &
    \includegraphics[width=0.125\linewidth]{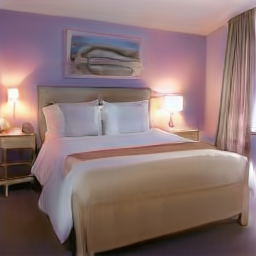} &
    \includegraphics[width=0.125\linewidth]{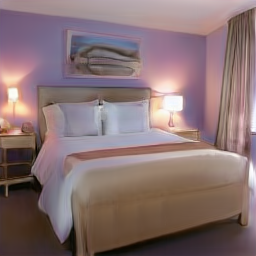} &
    \includegraphics[width=0.125\linewidth]{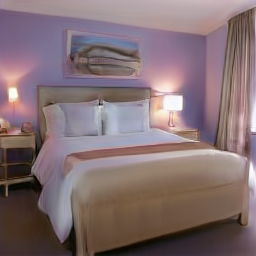} &
    \includegraphics[width=0.125\linewidth]{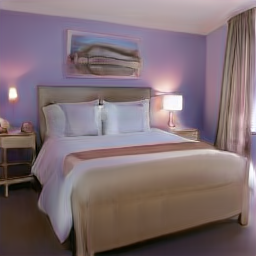} &
    \includegraphics[width=0.125\linewidth]{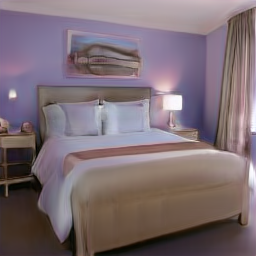} &
    \includegraphics[width=0.125\linewidth]{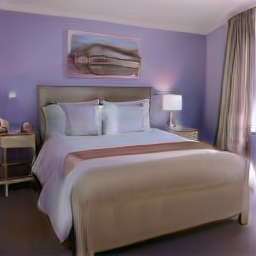} 
        \\
    \vspace{-2pt}
    \end{tabular}
    \centering
    \vspace{-5pt}
\includegraphics[width=1.01\linewidth]{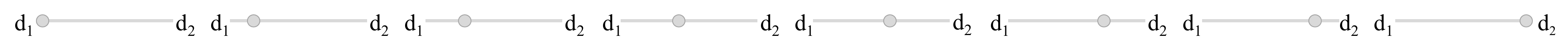}
\caption{Controllability: The first and last columns of the figure show relighted images generated using our Relit-1 and Relit-5 directions. The bottom section of the figure features a user-controllable slider that enables adjusting the weight of the relighting effects produced by these two directions. Moving the slider from left to right results in a seamless interpolation between the two lighting directions, providing users with precise control over the relighting of the generated images.}
    \label{interpolation_transient}
    \vspace{-10pt}
\end{figure*}

\begin{figure*}[t]
  \footnotesize
  \setlength\tabcolsep{0.1pt}
  \renewcommand{\arraystretch}{0.1}
  \begin{tabular}{cccccccc}
    \includegraphics[width=0.125\linewidth]{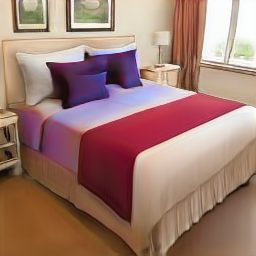} &
    \includegraphics[width=0.125\linewidth]{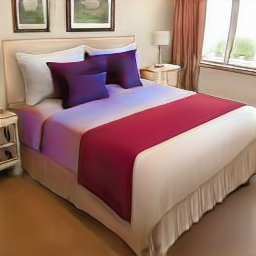} &
    \includegraphics[width=0.125\linewidth]{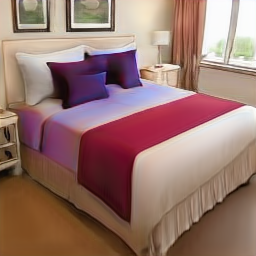} &
    \includegraphics[width=0.125\linewidth]{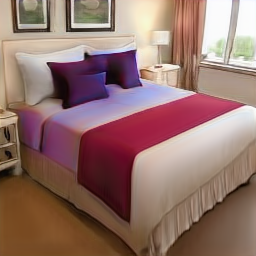} &
    \includegraphics[width=0.125\linewidth]{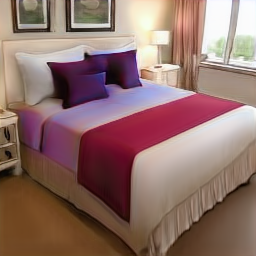} &
    \includegraphics[width=0.125\linewidth]{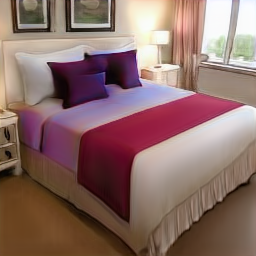} &
    \includegraphics[width=0.125\linewidth]{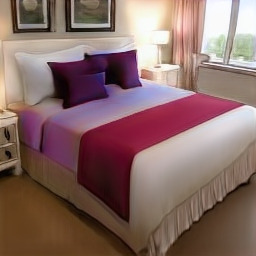} &
    \includegraphics[width=0.125\linewidth]{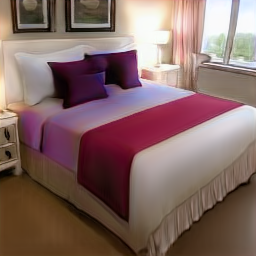} 
        \\
    \includegraphics[width=0.125\linewidth]{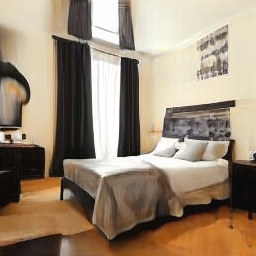} &
    \includegraphics[width=0.125\linewidth]{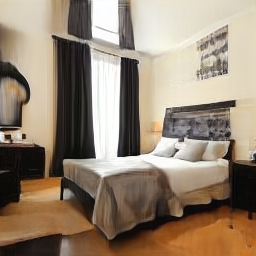} &
    \includegraphics[width=0.125\linewidth]{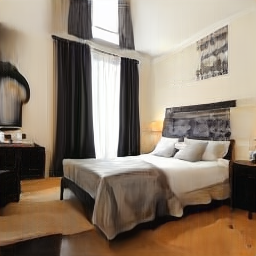} &
    \includegraphics[width=0.125\linewidth]{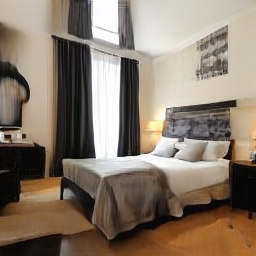} &
    \includegraphics[width=0.125\linewidth]{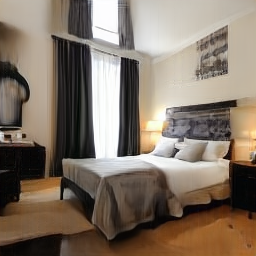} &
    \includegraphics[width=0.125\linewidth]{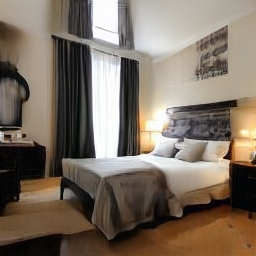} &
    \includegraphics[width=0.125\linewidth]{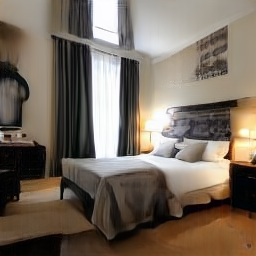} &
    \includegraphics[width=0.125\linewidth]{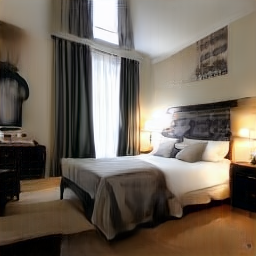} 
        \\
    \vspace{-2pt}
    \end{tabular}
    \centering
    \vspace{-7pt}
\hspace{-5pt}\includegraphics[width=1.02\linewidth]{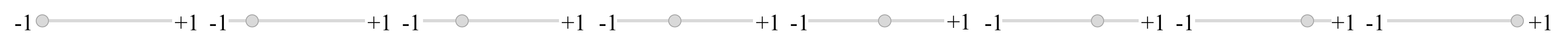}
\caption{Scaling Directions:  The figure depicts the persistent and smooth effects of applying the direction at different scalar coefficients. We use our Relit-2 direction. A slider at the bottom allows the user to adjust the weight of the relighting direction, producing a seamless interpolation when increasing or decreasing the intensity of the chosen direction. The relighting effects range from a well-lit room with the bedside lamp off to weak external lighting with a bedside lamp on.} 
    \label{interpolation_scale}
    \vspace{-10pt}
\end{figure*}

\begin{figure}[t!]
  \centering
  \includegraphics[width=0.975\linewidth]{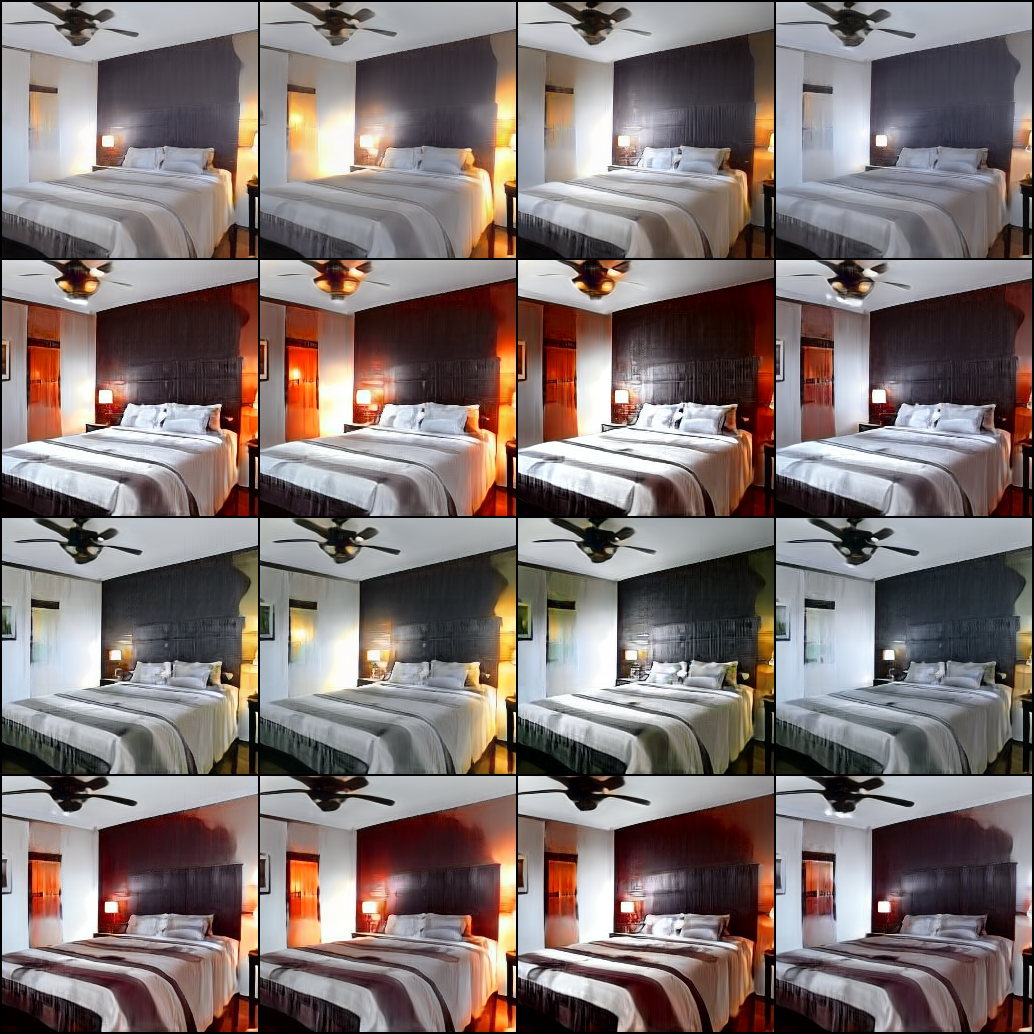}
 \caption{Simultaneous Joint Relighting and Resurfacing. {\bf Columns} show different relights for a fixed resurfacing, and {\bf rows} show different resurfacing for a fixed relight. The interactions between relighting and resurfacing are largely disentangled. 
  \label{both}}
  \vspace{-12pt}
\end{figure}

\section{Experiments}

\begin{figure}[t]
  \centering
\scriptsize
  \centering
  \footnotesize
  \setlength\tabcolsep{0.2pt}
  \renewcommand{\arraystretch}{0.1}

  \begin{tabular}{cccc}
\multicolumn{1}{c}{\rotatebox{90}{\hspace{20pt} Real}}  &     \includegraphics[width=0.28\linewidth, height=0.28\linewidth]{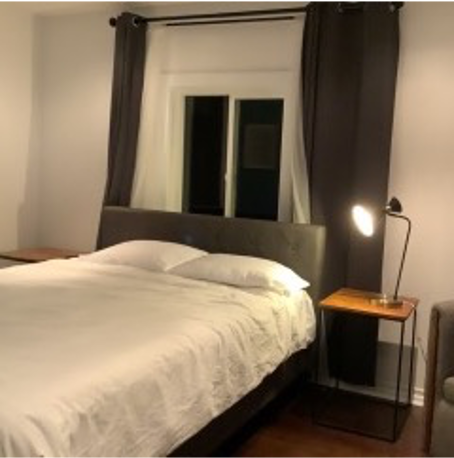} & \includegraphics[width=0.28\linewidth, height=0.28\linewidth]{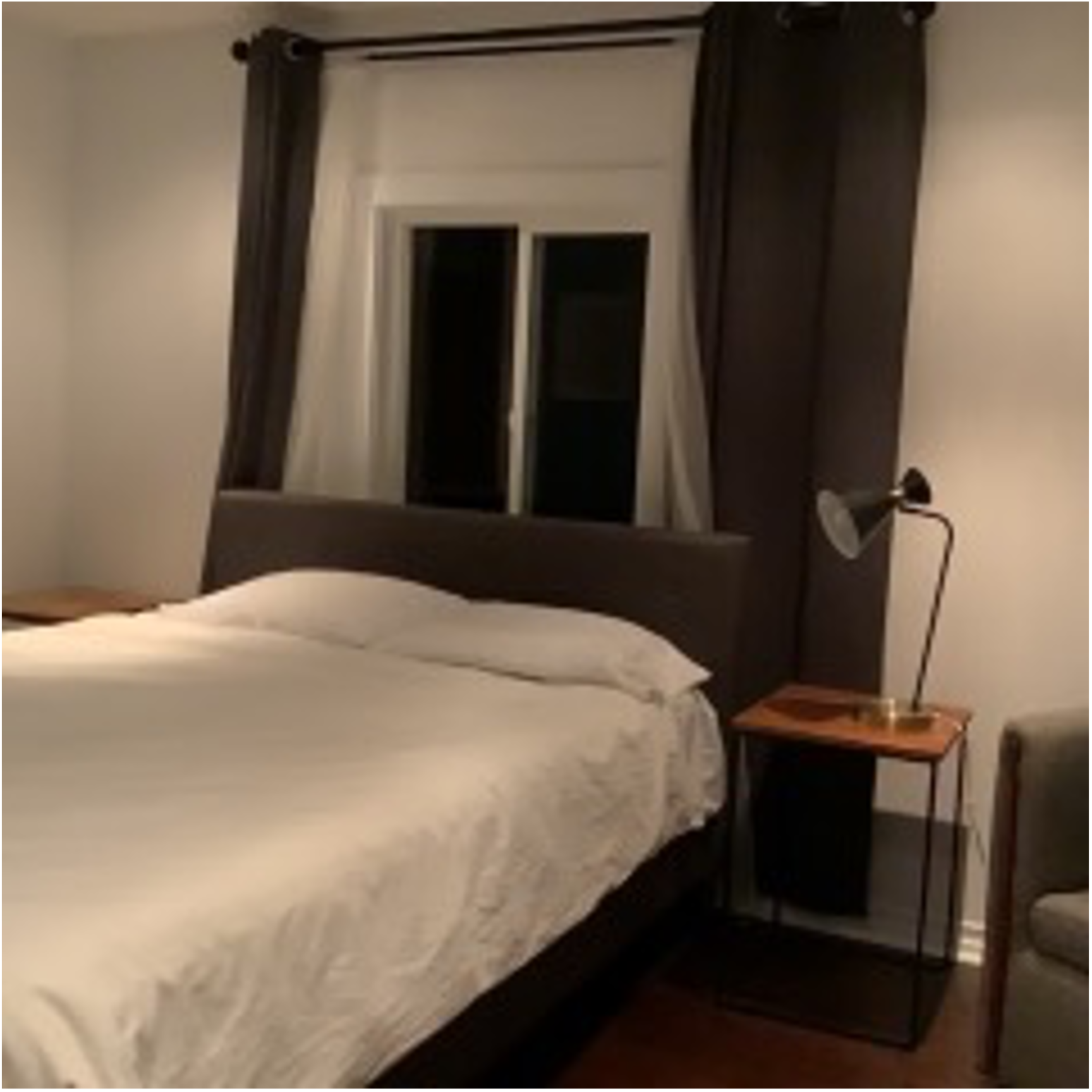} 
&\includegraphics[width=0.28\linewidth, height=0.28\linewidth]{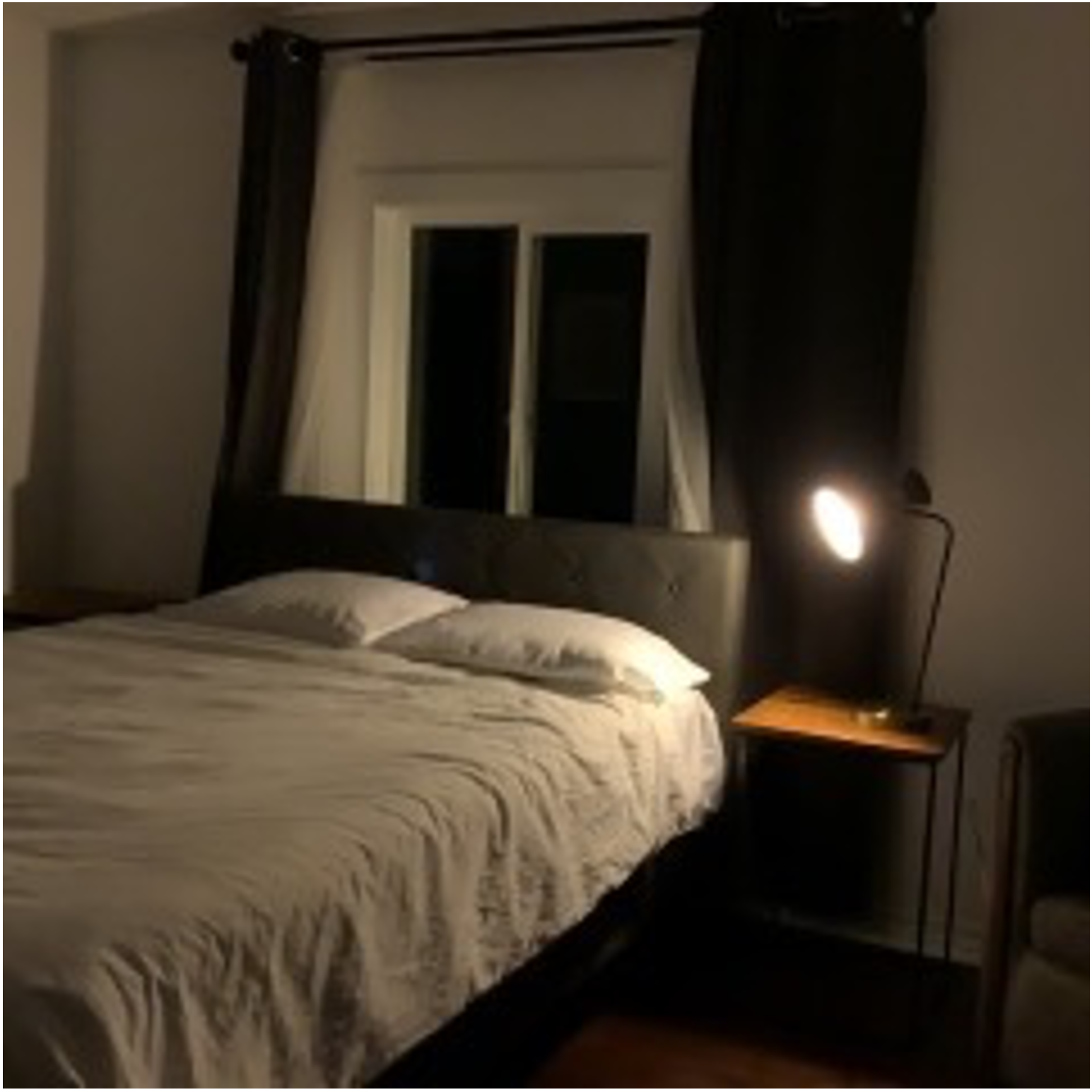}
    \\
\multicolumn{1}{c}{\centering \rotatebox{90}{\hspace{20pt} Li et al.}}  &      & \includegraphics[width=0.28\linewidth, height=0.28\linewidth]{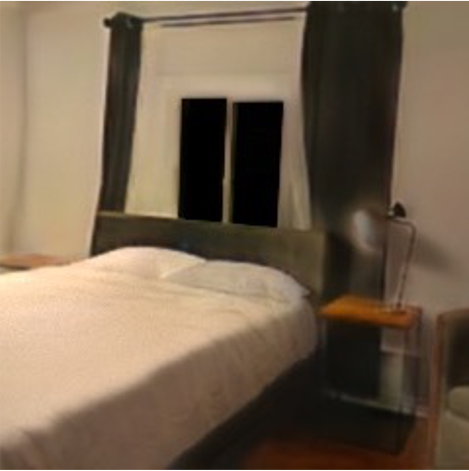} 
&\includegraphics[width=0.28\linewidth, height=0.28\linewidth]{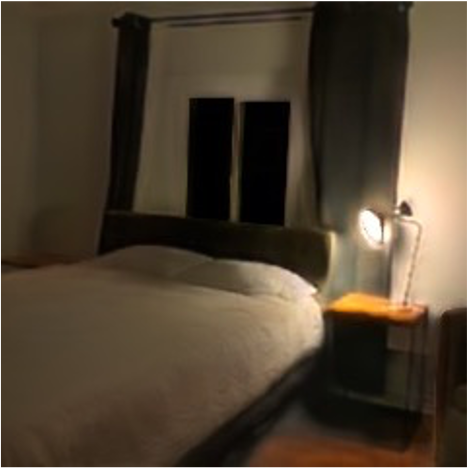}
    \\
\multicolumn{1}{c}{\rotatebox{90}{\hspace{20pt} Ours}}  &     \includegraphics[width=0.28\linewidth, height=0.28\linewidth]{figures/inverse_render_compare/real.png} & \includegraphics[width=0.28\linewidth, height=0.28\linewidth]{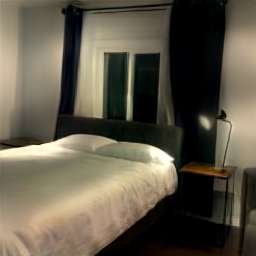} 
&\includegraphics[width=0.28\linewidth, height=0.28\linewidth]{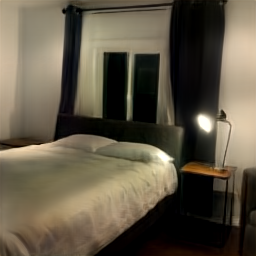}
    \\

   &  Inverted & Visible Light Off &  Invisible Light Off 
        \end{tabular}
\vspace{3pt}
\caption{With an appropriate inversion method~\cite{bhattad2023make}, we can relight real scenes.  {\bf Top row:} shows
  three images of the same scene in different lightings, obtained from~\cite{li2022physically}.
  {\bf Middle row:} shows relightings of the first image, obtained by Li {\em et al.} by inverse rendering, changing
  luminaire parameters, then forward rendering~\cite{li2022physically}.  {\bf Bottom left:} image generated by passing
  latent variables from our inversion of the top left through StyleGAN.  {\bf Bottom center} and {\bf bottom right} show
  relightings obtained by adding a relighting direction to these latent variables. Recall the relighting directions are
  {\em independent of image}.  The directions were selected by hand to
  correspond to the top center, right respectively; directions are Relit - 5 (visible light off) and Relit - 2
  (invisible light off) from Figure~\ref{fig:teaser}.  Note: our relights compare well to real images; our relights
  do not ring on fine edges (eg the lamp); our relights preserve high spatial frequencies in the image; and do not require CGI, physical rendering, or light source annotation.} 
      \label{fig:li2022comparison}
\vspace{-15pt}
\end{figure}

{\noindent \bf Qualitative evaluation:}  
StyLitGAN produces realistic images, that are out of distribution but known to exist for
straightforward physical reasons. Because they're out of distribution, current quantitative evaluation tools do not
apply. We evaluate realism qualitatively. Further, there is no direct comparable method. However, we show relighting
comparisons to a recent SOTA method that is physically motivated and trained with CGI data~\cite{li2022physically}
(Figure~\ref{fig:li2022comparison}). For  relighting, our method should generate images that: are clearly relightings of
a scene; fix geometry and albedo but visibly change shading; and display complicated illumination effects, including
soft shadows, cast shadows 
and gloss. 
For resurfacing, our method should generate images that: are clearly images of the original layout, but with different
materials or colors or changes in furniture; 
and display illumination effects that are consistent with these color
changes. As Figure~\ref{resurfacing} shows, our method meets these goals. 
Figure~\ref{interpolation_transient} and Figure~\ref{interpolation_scale} show  interpolation sequences for a
relighting between two directions and scaling only one direction.  Note that the lighting changes smoothly, as one would expect. Figure~\ref{both} shows our relighting and recoloring directions are largely disentangled. 

{\bf  Quantitative evaluation:}  Figure~\ref{fig:model_selection} shows how we can evaluate albedo shift and lighting shift. Further, we show we can generate image datasets with {\em increased} FID~\cite{parmar2021cleanfid, heusel2017gans}  compared to the base
comparison set in Table~\ref{fid-table} (we use clean-FID~\cite{parmar2021cleanfid}). This is strong evidence our method can produce a set of images that is a strict superset of those
that the vanilla StyleGAN can produce.

{\noindent \bf Generality:} We have applied our method to StyleGAN instances trained on different datasets
-- Conference Room, Kitchen, Living Room, Dining Room and Church and Faces (results in supplementary;
relighting is successful for each).  
\begin{table}[t!]
  \caption{FID measures distribution shift and not realism. Our generated images are realistic and are out-of-distribution because of large illumination and color changes in the images. This results in large FID scores.  $KDL$ in the table is for kitchen, dining and living room which are jointly trained ~\cite{epstein2022blobgan}.
  }
  \label{fid-table}
  \centering
  \resizebox{1\linewidth}{!}{%
  \begin{tabular}{llllll}
    \toprule
    Type     & Bedroom  & KDL & Conference & Church & Faces \\
    \midrule
    StyleGAN (SG) & 5.01 & 5.86  & 9.35 & 3.80 & 5.02 \\
    SG + Relighting (RL)
    & 14.23  & 6.87 & 10.48 &  12.12 & 37.87  \\
    SG + Resurfacing (RS
    &    17.03   & 9.41 & 10.63 & 18.60 & 34.06 \\
    SG + RL + RS
    &  21.39 & 11.68  & 12.71 & 21.08 & 37.40 \\ 
    \bottomrule
  \end{tabular}
  }
  \vspace{-10pt}
  \end{table}

\section{Discussion}
\label{discussion}

StyleGAN's relighting suggests it ``knows'' quite a lot about images.  It knows where bedside lights and windows are. 
Our method can't produce targeted relighting, and we can't guarantee that we have discovered all the images that StyleGAN can't generate by our prompting. We found new images by imposing a simple, necessary property of the distribution of images (scenes have many different lightings), and we expect that similar observations may lead to other types of out-of-distribution images.
Our relighting and resurfacing quality is dependent on the quality of the generative model. If StyleGAN is unable to produce realistic images, then our relighting results will also be unrealistic. All other limitations of StyleGAN also apply to our approach.

\section*{Acknowledgment}
We thank Aniruddha Kembhavi, Derek Hoiem, Min Jin Chong, and Shenlong Wang for their feedback and suggestions. This material is based upon work supported by the National Science Foundation under Grant No. 2106825 and by a gift from the Boeing Corporation. 

{\small
\bibliographystyle{ieee_fullname}
\bibliography{main}
}

\appendix

\newpage
\section*{\large Supplementary Material for StyLitGAN}
\section{Choice of Decomposition}
\begin{figure}[htpb!]
  \centering
  \includegraphics[width=1\linewidth]{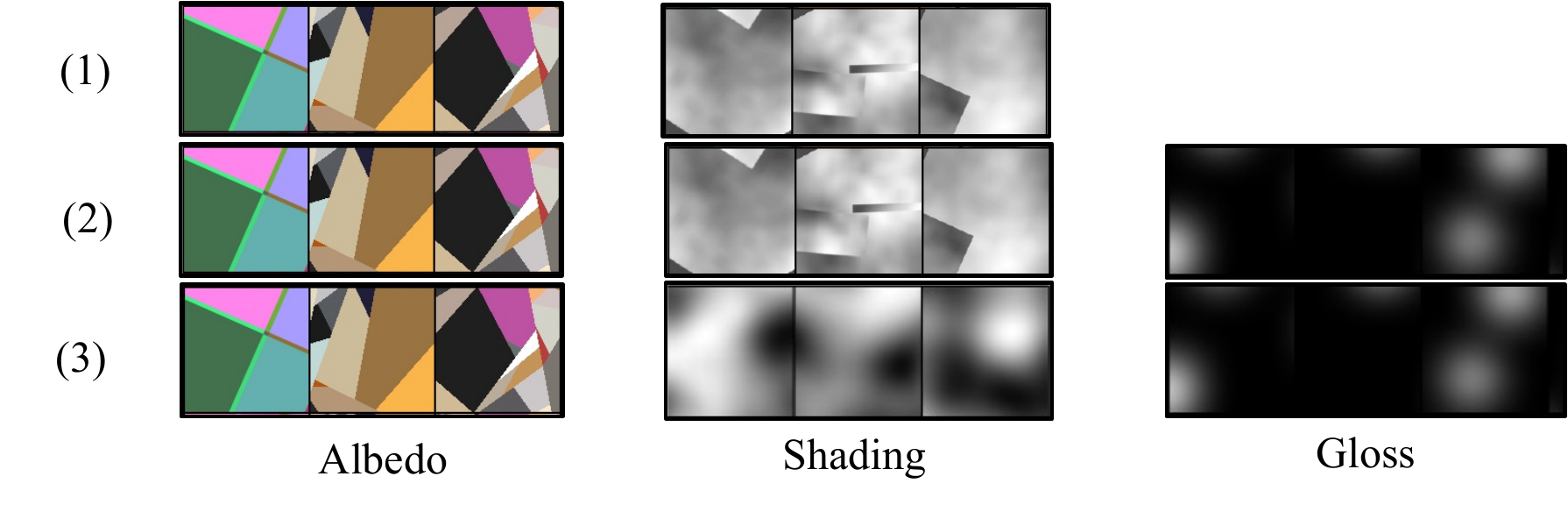}
  \caption{We compare three different decomposition models derived from~\cite{forsyth2020intrinsic}. We find decomposition (3) to provide significantly better results with small geometry shifts after relighting. In general, the directions obtained with this decomposition are admissible. The other two decompositions resulted in relighting directions with large albedo changes. The major differences between these decomposition methods are (a) we use an additional gloss component in (2) and (3) and (b) we also assume shading to be smooth and have all the high-frequency fine-edges information in the albedo to preserve geometric details without noticeable shifts after relighting in (3).}
  \label{our_decomp}
  \vspace{-10pt}
\end{figure}

The choice of decomposition matters for relighting without change in geometry and albedo. The best-performing decomposition that was admissible from our experiments has been a variant decomposition that models fine edges in albedo rather than in the shading field. 
As we apply diversity loss on the shading field; it is practical to not model geometry (fine edges; normals). Otherwise, undesirable geometry shifts may occur, as demonstrated in the videos on our project page. Representative examples of our modified decomposition can be found in Fig.~\ref{our_decomp}. Furthermore, we observed that incorporating gloss as an additional component enhances the identification of light sources and facilitates more realistic lighting alterations while maintaining diverse appearance changes.

\section{Albedo Scores and Analogy with CCA}
The relevant maximum can be computed by analogy with canonical correlation analysis.
Reshape each color component of the patch into an $(3 M N)$ vector $\vect{P}$ with components
$p_{w(i,j), k}$. Construct $(M N)$basis vectors $x_{w(i, j)}=i$ and $y_{w(i, j)}=j$.  Now construct
the $(3 M N) \times 3$ matrix ${\cal M}(\vect{p})$ so that
\begin{equation}
\small 
\vect{P}'(q_x, q_y, q_c)= {\cal M}(\vect{p})\vect{q}_c =\left[\begin{array}{ccc}
                             x_1 p_{1,1}& y_1 p_{1, 1}& p_{1, 1}\\
                             \ldots&\ldots&\ldots\\
                             x_1 p_{1, 2}&y_1 p_{1, 2}& p_{1, 2}\\
                             \ldots&\ldots&\ldots\\
                             x_1 p_{1, 3}&y_1 p_{1, 3}& p_{1, 3}\\
                             \ldots&\ldots&\ldots
                           \end{array}\right]\left[\begin{array}{c} q_x\\q_y\\q_c\end{array}\right]          
\end{equation}
so that  $d_a(\vect{p}, \vect{q}) $
\begin{align}
d_a(\mathbf{p}, \mathbf{q}) &= 1 - \frac{\dafmax{\mathbf{p}_c, \mathbf{q}_c}(\mathbf{q}_c \mathbf{M}'(\mathbf{p})\mathbf{M}(\mathbf{q})\mathbf{p}_c)}{\sqrt{(\mathbf{q}_c \mathbf{M}'(\mathbf{p})\mathbf{M}(\mathbf{p})\mathbf{q}_c)(\mathbf{p}_c \mathbf{M}'(\mathbf{q})\mathbf{M}(\mathbf{q})\mathbf{p}_c)}} \\
&\hphantom{=} 1 - \frac{\dafmax{\mathbf{p}_c,\mathbf{q}_c} \mathbf{q}_c \Sigma_{xy}\mathbf{p}_c}{\sqrt{(\mathbf{q}_c \Sigma_{xx}\mathbf{q}_c)(\mathbf{p}_c \Sigma_{yy} \mathbf{p}_c)}}.
\end{align}

  Standard results then yield that
\begin{equation}
  d_a(\vect{p}, \vect{q})=1-\sqrt{\lambda_x}
\end{equation}
where $\lambda_x$ is the largest eigenvalue of
\begin{equation}
  \Sigma_{xx}^{-1}\Sigma_{xy}  \Sigma_{yy}^{-1}\Sigma_{xy}^T
  \end{equation}

\section{Additional Qualitative Examples and Movies}

For better visualization, we provide interpolation movies on our project page. We use a simple linear interpolation between distinct relighting directions that we found. The movies show smooth continuous lighting changes with very small local geometry changes.

\section{Other Experimental Details}

For our Model 14 relighting, we employ the following $\lambda$ coefficients: $\lambda_{const} = 750, \lambda_{per} = 0.1, \lambda_{dist} = 1, \lambda_{deco} = 0.01$. We also apply distinct $\lambda_{div}$ values for different categories. For bedrooms, we use $\lambda_{div} = 0.125$; for kitchens, dining, and living rooms, $\lambda_{div} = 0.25$; for conference rooms, $\lambda_{div} = 0.4$; for faces, $\lambda_{div} = 0.1$; and for churches, $\lambda_{div} = 0.5$. It is important to note that these coefficients pertain to the selected model with albedo, shading, and gloss decomposition, and fine edges are modeled in albedo, as previously discussed.

For our recoloring or resurfacing, we use the following $\lambda$ coefficients: $\lambda_{const} = 1000, \lambda_{per} = 0.1, \lambda_{dist} = 1, \lambda_{deco} = 0$. We also employ different $\lambda_{div}$ values for various categories. For bedrooms, we use $\lambda_{div} = 0.3$; for kitchens, dining, and living rooms, $\lambda_{div} = 1$; for conference rooms, $\lambda_{div} = 0.5$; for faces, $\lambda_{div} = 0.2$; and for churches, $\lambda_{div} = 0.6$.

For all categories, we employ 2000 search iterations; however, effective relighting directions become apparent after only a few hundred iterations. In addition, we utilize the Adam optimizer for searching the latent directions with a learning rate of 0.001 and for updating the classifier with a learning rate of 0.0001.

\end{document}